\documentclass{article}

\usepackage[main,final]{neurips_2026}

\usepackage[utf8]{inputenc}
\usepackage[T1]{fontenc}
\usepackage{hyperref}
\usepackage{url}
\usepackage{booktabs}
\usepackage{placeins}
\usepackage{enumitem}
\usepackage{nicefrac}
\usepackage{microtype}
\usepackage{xcolor}

\usepackage{amsfonts}
\usepackage{amssymb}
\usepackage{amsmath}
\usepackage{amsthm}
\usepackage{mathtools}
\usepackage{bm}

\usepackage{algorithm}
\usepackage{algpseudocode}

\DeclareRobustCommand{\GBC}{\ifmmode\mathrm{GBC}\else GBC\fi}

\newcommand{\non}{n_{\mathrm{on}}}
\newcommand{\noff}{n_{\mathrm{off}}}
\newcommand{\Don}{\mathcal{D}_{\mathrm{on}}}
\newcommand{\Doff}{\mathcal{D}_{\mathrm{off}}}

\newcommand{\vhat}{\hat{v}}
\newcommand{\ehat}{\hat{e}}
\newcommand{\vzero}{v_0}
\newcommand{\vzeroGBC}{\hat{v}_0^{\GBC}}
\newcommand{\vzeroGBCB}{\hat{v}_{0,B}^{\GBC}}
\newcommand{\vK}{\hat{v}^{(K)}}

\newcommand{\Tpi}{T_\pi}
\newcommand{\Ppi}{P_\pi}
\newcommand{\Tpihat}{\widehat{T}_\pi}
\newcommand{\Tzeropihat}{\widehat{T}_{0,\pi}}
\newcommand{\TGBC}{T_\pi^{\GBC}}
\newcommand{\TGBCB}{T_\pi^{\GBC,B}}
\newcommand{\PpiE}{P_{\pi,\ehat}}
\newcommand{\PpiEB}{P_{\pi,\ehat,B}}

\newcommand{\PiE}{\Pi_{\ehat}}
\newcommand{\PiEB}{\Pi_{\ehat,B}}
\newcommand{\PiV}{\Pi_{\vhat}}

\newcommand{\FB}{\mathcal{F}_B}
\newcommand{\FBE}{\mathcal{F}_{B,\ehat}}
\newcommand{\FE}{\mathcal{F}_{\ehat}}

\newcommand{\CalGBCB}{\mathrm{Cal}^{\GBC}_B}

\newcommand{\wpi}{w_\pi}
\newcommand{\wpihat}{\widehat{w}_\pi}

\newcommand{\E}{\mathbb{E}}

\newcommand{\Cadapt}{C_{\mathrm{rate}}}
\newcommand{\siglab}{\sigma}
\newcommand{\estar}{\varepsilon_{\mathrm{approx}}}
\newcommand{\BEhat}{\varepsilon_{\mathrm{BE},\ehat}}
\newcommand{\Cocc}{C_{\mathrm{occ}}}
\newcommand{\CTV}{C_{\mathrm{TV}}}

\newcommand{\termRB}{(\mathrm{R}_B)}
\newcommand{\termR}{(\mathrm{R})}
\newcommand{\termRMC}{(\mathrm{R\text{-}MC})}
\newcommand{\termRTD}{(\mathrm{R\text{-}TD})}
\newcommand{\termW}{(\mathrm{W})}
\newcommand{\termI}{(\mathrm{I})}
\newcommand{\termS}{(\mathrm{S})}
\newcommand{\termN}{(\mathrm{N})}

\newtheorem{theorem}{Theorem}
\newtheorem{proposition}[theorem]{Proposition}

\newtheorem{lemma}[theorem]{Lemma}
\theoremstyle{definition}
\newtheorem{definition}[theorem]{Definition}
\newtheorem{assumption}[theorem]{Assumption}
\theoremstyle{remark}
\newtheorem{remark}[theorem]{Remark}

\title{Error-Guided Bellman Calibration\\for Semi-Offline Value Estimation}

\author{%
  Tetsuro Morimura\\
  CyberAgent\\
  Tokyo, Japan\\
  \texttt{morimura\_tetsuro@cyberagent.co.jp}\\
}

\begin{document}

\maketitle

\begin{abstract}
We study semi-offline value estimation, where a small on-policy sample supplements a large off-policy dataset.
This regime is common in recommender systems and pre-deployment A/B testing, and the challenge is to use the limited on-policy data to correct systematic bias in a given value predictor.
Recent work on Bellman calibration (BC) gives a post-hoc, model-agnostic correction along the predicted value.
We propose Error-Guided Bellman Calibration (GBC), which trains an error model on the on-policy sample and calibrates the predictor along this learned error axis.
We bound the refinement term, which measures the part of the prediction error that calibration along this axis cannot remove, in terms of the accuracy of the error model.
Under low-dimensional structure and an assumed error-model rate, we show that this bound adapts to the intrinsic dimension rather than the state dimension, and we give a finite-sample bound in which this structure enters only through the refinement term.
On a synthetic MDP, a customer relationship management simulator, and D4RL tasks, GBC reliably reduces the error of the given predictor and consistently improves on BC.
\end{abstract}

\section{Introduction}\label{sec:intro}

Off-policy value estimation underlies a large class of decisions in modern machine-learning systems, from podcast and advertising recommenders \citep{theocharous2015lifetime,hallak2017ltv,maystre2022spotify} to industrial counterfactual reasoning \citep{bottou2013counterfactual} and pre-deployment policy comparison \citep{precup2000eligibility,levine2020offline,uehara2022opereview}. The standard pipeline trains a value predictor on logged data and uses its output to score candidate policies. In practice, the predictor is systematically biased. Sources of this bias include distribution shift between the logging and target policies, model misspecification, and incomplete training. Their combined effect is hard to remove by retraining alone. Calibration offers a post-hoc alternative. It keeps the trained predictor and adjusts its outputs using additional data.

Many applications produce data that is neither fully on-policy nor fully off-policy. A large logged dataset accumulates from past deployments and historical A/B tests, while a much smaller on-policy sample is collected under the target policy. Deployment cost and the risk of running an untested policy keep the on-policy budget small, even when the off-policy log is abundant \citep{levine2020offline}. We call this setting the \emph{semi-offline} regime. Hybrid reinforcement learning combines the two sources for policy learning \citep{song2023hybrid}; we study how to combine them to correct a given value predictor.

A natural starting point is the recently proposed Bellman calibration\footnote{We use BC for Bellman Calibration throughout; behavior cloning does not appear in this paper.} (BC) of \citet{vanderlaan2025bc}. BC takes a value predictor $\vhat$ as input and applies a histogram calibrator along the level sets of $\vhat$, enforcing a one-step Bellman consistency within each bin via a doubly robust target. The procedure is post-hoc and model-agnostic, and it avoids both Bellman completeness and realizability assumptions on $\vhat$. \citet{vanderlaan2025bc} also note that the calibration step can be run on a limited amount of on-policy data. We adopt BC's framework as the foundation: the doubly robust Bellman target, iterated histogram calibration, the coarsened Bellman operator, and the calibration-refinement decomposition \citep{degroot1983comparison} with its finite-sample analysis.

A histogram calibrator groups states by a scalar variable and applies one adjustment to each group. We call this variable the conditioning axis. BC uses the prediction $\vhat$ itself as the conditioning axis, so states with the same prediction receive the same calibrated value, even when their errors differ. Let $\vzero$ be the true value function, and define the base prediction error as $e := \vzero - \vhat$, so that $\vzero = \vhat + e$. Calibration along $\vhat$ cannot remove the part of $e$ that varies across states with the same prediction. In BC's bound on the estimation error, the size of this part appears as a separate term, the refinement term. This part can be large even when $e$ has a simple form. For example, let $\vzero = u + g \circ \phi$, where $(g \circ \phi)(s) = g(\phi(s))$. If the base predictor captures only the first component, $\vhat = u$, then $e = g \circ \phi$ depends on the state only through $\phi$. Yet when $u(S)$ and $\phi(S)$ are independent, BC's grouping by $u$ reveals nothing about $\phi$, so calibration along $\vhat$ cannot correct any of the variation in $g \circ \phi$ (Remark~\ref{rem:bc-failure}). A model of the prediction error, rather than of the value, can distinguish states that share a prediction but differ in $e$.

We propose Error-Guided Bellman Calibration (\GBC{}). Rather than adding the small on-policy sample to the calibration data, \GBC{} trains an error model $\ehat$ on it. It then uses $\ehat$ as the conditioning axis and calibrates $\vhat$ on an off-policy calibration sample with the additive update $\vhat + \theta \circ \ehat$. \GBC{} keeps $\vhat$ and lets the correction vary with $\ehat$, so states with the same $\vhat$ but different $\ehat$ can receive different corrections. Because the correction is additive, it can remove the part of $e$ that is predictable from $\ehat$. When $\ehat$ tracks $e$, this part accounts for most of $e$ (Proposition~\ref{prop:mc-refinement}).

Our contributions are threefold.
\begin{itemize}[leftmargin=*]
\item \textbf{Algorithmic.} \GBC{} learns an error axis from the on-policy sample and calibrates $\vhat$ along it with an additive correction (Section~\ref{sec:gbc}).
\item \textbf{Theoretical.} We prove bounds on the refinement term in terms of the accuracy of $\ehat$ (Propositions~\ref{prop:mc-refinement} and~\ref{prop:td1-refinement}). Under low-dimensional structure and an assumed error-model rate, these bounds adapt to the intrinsic dimension rather than the state dimension (Theorems~\ref{thm:rate-transfer} and~\ref{cor:rate-transfer-td1}). We also give a finite-sample bound, built on BC's analysis, in which this structure enters through the refinement term (Theorem~\ref{thm:combined}).
\item \textbf{Empirical.} We evaluate \GBC{} on a controlled synthetic toy MDP, the customer relationship management (CRM) benchmark of \citet{vanderlaan2025bc}, and three D4RL continuous-control tasks \citep{fu2020d4rl}. In every environment, \GBC{} improves on the uncalibrated predictor, even where BC makes it worse. \GBC{} also consistently improves on BC (Section~\ref{sec:experiments}).
\end{itemize}

\paragraph{Related work.}
The closest predecessor is BC \citep{vanderlaan2025bc}; Section~\ref{sec:gbc-vs-bc} compares the two methods. BC extends the calibration-with-nuisance program of \citet{vanderlaan2023ic} and \citet{whitehouse2024orthogonal} to MDPs. Iterated calibration in MDPs also appears in \citet{vanderlaan2025sdrl}, who use it to debias plug-in estimators of long-term policy values. A companion paper of \citet{vanderlaan2025swfqe} instead changes how the predictor is trained, reweighting the objective of fitted Q evaluation \citep{le2019batch} with a stationary density ratio. This approach is complementary to ours, which leaves the predictor fixed and adjusts its outputs. Hybrid RL \citep{song2023hybrid,zhou2023offlinepg} shares the small-on-policy plus large-off-policy structure but targets policy learning rather than value estimation. Off-dynamics RL \citep{eysenbach2021offdyn} addresses dynamics shift across domains rather than bias within a single MDP. Appendix~\ref{app:related} provides an extended discussion.

\section{Setup}\label{sec:setup}

\paragraph{MDP and value function.}
We work in the standard infinite-horizon discounted Markov decision process (MDP) with state space $\mathcal{S}$, finite action space $\mathcal{A}$, transition kernel $P(s' \mid s, a)$, bounded reward function $r(s, a)$, and discount $\gamma \in [0, 1)$, largely following the notation of \citet{vanderlaan2025bc}. The target policy $\pi$ is known; the behavior policy $b$ that generated the off-policy data (described below) may be known or unknown. The target policy induces the Bellman operator $\Tpi v := r_\pi + \gamma \Ppi v$, where $r_\pi(s) := \sum_a \pi(a \mid s)\, r(s, a)$ and $(\Ppi v)(s) := \E[v(S') \mid S = s, A \sim \pi(\cdot \mid s)]$. Its unique bounded fixed point $\vzero = \Tpi \vzero$ is the true value function under the target policy $\pi$,
\[
    \vzero(s) = \E_\pi\Big[\sum_{t \ge 0} \gamma^t r(S_t, A_t) \,\Big|\, S_0 = s\Big].
\]
The estimand is $\vzero$.

\paragraph{Data and base predictor.}
The data setup is semi-offline, with two data sources. A small set $\Don$ of on-policy transitions $(S, A, R, S')$ is collected by running $\pi$, with $\non := |\Don|$. Observed rewards satisfy $R = r(S, A) + \varepsilon$ with $\E[\varepsilon \mid S, A] = 0$. A large off-policy calibration sample $\Doff$ of transitions $(S, A, R, S')$ is collected under $b$ and used to fit the calibrator, with $\noff := |\Doff|$. The regime of interest is $\non \ll \noff$. The base predictor $\vhat: \mathcal{S} \to \mathbb{R}$ is an arbitrary bounded value estimator supplied as input to the algorithm. Our analysis assumes that $\Don$, $\Doff$, and any data used to fit $\vhat$ are mutually independent. In each transition of $\Doff$, the state $S$ is drawn from a distribution $\rho$. For a state distribution $\nu$, $\E_\nu$ denotes expectation over $S \sim \nu$. For a function ${h : \mathcal{S} \to \mathbb{R}}$, we write $\|h\|_\nu := \E_\nu[h(S)^2]^{1/2}$ for its $L^2(\nu)$ norm.

\paragraph{Doubly robust target.}
To target $\Tpi$ from off-policy transitions, we follow BC \citep{vanderlaan2025bc} and use the doubly robust (DR) Bellman target, standard in off-policy evaluation \citep{dudik2011doubly,jiang2016dr}. It combines importance weighting with a fitted reward-and-transition model. It is unbiased if either is correct, its bias is otherwise the product of their errors, and the model can reduce the variance of importance weighting. The importance ratio $\wpi := \pi/b$ is computed directly when $b$ is known; otherwise, $b$ (or the ratio itself) is estimated. We denote the resulting weights, possibly clipped, by $\wpihat$. Given $\wpihat$ and nuisance estimates $\hat{r}$ of $r$ and $\hat{P}$ of $P$, set $\hat{q}_v(s, a) := \hat{r}(s, a) + \gamma (\hat{P} v)(s, a)$ and define the observation-level operator
\begin{equation}\label{eq:dr-bellman}
    \Tpihat(v)(S, A, R, S') := \sum_a \pi(a \mid S)\,\hat{q}_v(S, a) + \wpihat(A \mid S)\,[R + \gamma v(S') - \hat{q}_v(S, A)].
\end{equation}
The analysis requires action overlap, $b(a \mid s) \ge \pi(a \mid s)/M$ for all $(s, a)$, so that every action taken by $\pi$ is also taken by $b$, and bounded weights $\wpihat$ (condition C1 in Appendix~\ref{app:thm2}).

\paragraph{Bellman calibration.}
A predictor $v$ is Bellman calibrated \citep{vanderlaan2025bc} if $v(s) = \E_\rho[\Tpi(v)(S) \mid v(S) = v(s)]$, that is, if its prediction equals the mean Bellman target among the states that share that prediction. The true value function $\vzero$ is Bellman calibrated because $\vzero = \Tpi \vzero$. A Bellman calibrated predictor need not equal $\vzero$, however, because the definition constrains only averages over states that share a prediction. The Bellman calibration (BC) procedure makes a given base predictor $\vhat$ approximately Bellman calibrated. As in histogram calibration \citep{gupta2021histogram}, it groups the states into $B$ bins by the value of $\vhat$ and gives all states in a bin the same new prediction, the mean of the DR target $\Tpihat(\vhat)$ over the transitions in the calibration sample whose state lies in that bin. The target is computed from the predictor being calibrated, so BC repeats this step for $K$ iterations, recomputing the target at each iteration from the latest predictions. Its output is constant on each bin and is therefore a function of $\vhat$ alone, written $\theta \circ \vhat$. We call the variable used to group the states the conditioning axis. For BC, it is $\vhat$.

\section{Error-Guided Bellman Calibration}\label{sec:gbc}

\subsection{Algorithm}\label{sec:gbc-algo}

Error-Guided Bellman Calibration (\GBC{}) takes a base predictor $\vhat$ as given and corrects it in two steps. First, it fits an error model $\ehat: \mathcal{S} \to \mathbb{R}$ from a model class $\mathcal{E}$ on the on-policy sample $\Don$, using the training targets described below, and holds $\ehat$ fixed thereafter. Second, it groups states into $B$ bins by the scalar value $\ehat(s)$ and runs $K$ iterations of histogram calibration on the calibration sample $\Doff$. Each iteration computes one correction per bin, the mean of the DR target minus $\vhat$ over the calibration transitions in that bin, and adds it to $\vhat$, giving the additive update $\vhat + \theta \circ \ehat$. Algorithm~\ref{alg:gbc} states the procedure, and Section~\ref{sec:gbc-vs-bc} compares it with BC.

Recall that $e := \vzero - \vhat$ is the prediction error of the base predictor. By default, $\ehat$ is fit by least squares to the training target $R_j + \gamma \vhat(S_j') - \vhat(S_j)$, the one-step temporal-difference (TD) error of the fixed base predictor on an on-policy transition $(S_j, A_j, R_j, S_j')$ in $\Don$. We call this TD training. The target's conditional mean given $S_j$ is $[(I - \gamma \Ppi) e](S_j)$, the Bellman residual of $\vhat$ at $S_j$, rather than $e(S_j)$. TD training needs only one-step transitions, and our experiments use it. Monte Carlo (MC) training instead uses the training target $G_j - \vhat(S_j)$, where $G_j$ is the discounted return from $S_j$; the target's conditional mean given $S_j$ is $e(S_j)$ itself. However, the return $G_j$ can be computed only when the rest of the episode from $S_j$ is observed without truncation. With either target, \GBC{} uses $\ehat$ only to group states, and Section~\ref{sec:refinement-training} analyzes both.

\begin{algorithm}[t]
\caption{Iterated \GBC{}}\label{alg:gbc}
\small{
\begin{algorithmic}[1]
\Require base predictor $\vhat$, on-policy sample $\Don$, error-model class $\mathcal{E}$, calibration sample $\Doff$, DR Bellman operator $\Tpihat$, histogram class $\FB$ with $B$ bins, iterations $K$
\State $\ehat \leftarrow \mathrm{TrainErrorModel}(\vhat, \Don, \mathcal{E})$ \Comment{fit error model on $\Don$}
\State $\hat{v}^{(0)} \leftarrow \vhat$
\For{$k = 0, 1, \ldots, K - 1$} \Comment{iterated calibration on $\Doff$}
  \State $\hat{\xi}_i^{(k)} \leftarrow \Tpihat(\hat{v}^{(k)})(O_i) - \vhat(S_i)$ \quad for each $O_i = (S_i, A_i, R_i, S_i') \in \Doff$
  \State $\theta_n^{(k+1)} \leftarrow \arg\min_{\theta \in \FB}\, \frac{1}{\noff} \sum_{i=1}^{\noff} \big(\hat{\xi}_i^{(k)} - \theta(\ehat(S_i))\big)^2$
  \State $\hat{v}^{(k+1)} \leftarrow \vhat + \theta_n^{(k+1)} \circ \ehat$
\EndFor
\State \Return $\vK$
\end{algorithmic}}
\end{algorithm}

The calibration loop is a one-dimensional histogram regression on $\ehat$. For a partition $I_1, \ldots, I_B$ of $\mathbb{R}$, $\FB$ is the class of functions that are constant on each $I_j$. For each calibration transition $O_i = (S_i, A_i, R_i, S_i')$ in $\Doff$, the pseudo-residual $\hat{\xi}_i^{(k)}$ compares $\vhat(S_i)$ with the DR Bellman target $\Tpihat(\hat{v}^{(k)})(O_i)$ defined in Eq.~\eqref{eq:dr-bellman}, and the least-squares step sets $\theta_n^{(k+1)}$ on each nonempty bin $I_j$ to the sample mean of $\hat{\xi}_i^{(k)}$ over the calibration transitions with $\ehat(S_i) \in I_j$. The bin count $B$ trades the approximation error within each bin against the number of samples per bin, a trade-off that Theorem~\ref{thm:combined} makes explicit. The analysis takes the partition as fixed, whereas our implementation places the bin edges at empirical quantiles of $\ehat$ (Appendix~\ref{app:impl}). Richer one-dimensional calibrators, such as isotonic regression, could replace $\FB$, but our guarantees cover the histogram class only.

\subsection{Comparison with BC}\label{sec:gbc-vs-bc}

\GBC{} adopts the DR Bellman target and the iterated histogram regression of BC (Section~\ref{sec:setup}). Unlike BC, it fits an error model $\ehat$, uses $\ehat$ instead of $\vhat$ as the conditioning axis, and adds a binwise correction to $\vhat$ instead of replacing the prediction. Section~\ref{sec:theory} shows how these differences enter the error bound through the refinement term.

\section{Theory}\label{sec:theory}

This section analyzes the estimation error of \GBC{}. We first decompose it into a calibration error and a refinement error (Section~\ref{sec:cal-ref-decomp}). Calibration reduces the first but cannot remove the second, which depends on the conditioning axis $\ehat$ and not on the calibration sample. We then bound the refinement error in terms of the accuracy of $\ehat$ (Section~\ref{sec:refinement-training}) and show that, under low-dimensional structure and an assumed convergence rate for $\ehat$, this bound adapts to the intrinsic dimension rather than the state dimension (Section~\ref{sec:lowdim-adaptation}). Finally, we give a finite-sample bound that combines the two components (Section~\ref{sec:finite-sample}). Proofs are deferred to Appendices~\ref{app:proofs} and~\ref{app:aux}.

\subsection{Calibration-Refinement Decomposition}\label{sec:cal-ref-decomp}

The population counterpart of one iteration of Algorithm~\ref{alg:gbc} replaces the DR target $\Tpihat$ by the true Bellman operator $\Tpi$ and the binwise sample means by conditional expectations given $\ehat$. For a bounded function $g: \mathcal{S} \to \mathbb{R}$, this gives the operator
\begin{equation}\label{eq:coarsened-op}
  \TGBC(g) := \vhat + \PiE[\Tpi(g) - \vhat],
\end{equation}
which we call the coarsened Bellman operator, following BC. Here $\PiE q$ denotes the $L^2(\rho)$ projection of $q$ onto functions of $\ehat$, that is, the function of $\ehat$ closest to $q$ in $\|\cdot\|_\rho$. It equals the conditional expectation
\[
  (\PiE q)(s) = \E_\rho[q(S) \mid \ehat(S) = \ehat(s)].
\]

\begin{lemma}[Coarsened Bellman fixed point]\label{lem:fixed-point}
The operator $\TGBC$ is a $\gamma$-contraction in $\|\cdot\|_\infty$ and therefore has a unique bounded fixed point $\vzeroGBC$. The difference $\vzeroGBC - \vhat$ is a function of $\ehat$.
\end{lemma}

Since $\vzeroGBC - \vhat$ is a function of $\ehat$, the fixed-point equation $\vzeroGBC = \TGBC(\vzeroGBC)$ and Eq.~\eqref{eq:coarsened-op} give $\PiE[\Tpi(\vzeroGBC) - \vzeroGBC] = 0$. That is, the Bellman residual of $\vzeroGBC$ has mean zero given $\ehat(S)$.

Let $\PpiE := \PiE\Ppi$ denote the coarsened transition operator. For a function ${h : \mathcal{S} \to \mathbb{R}}$,
\begin{equation}\label{eq:coarsened-trans}
  (\PpiE h)(s) = \big(\PiE(\Ppi h)\big)(s) = \E_\rho[(\Ppi h)(S) \mid \ehat(S) = \ehat(s)],
\end{equation}
the mean of $\Ppi h$ over the states $S$ with $\ehat(S) = \ehat(s)$. Like $\Ppi$, it is the transition operator of a Markov chain. We measure the estimation error under a stationary distribution of this chain and relate it to $\rho$ through a coverage coefficient.

\begin{assumption}[Stationary coverage]\label{ass:stat-cov}
The operator $\PpiE$ has a stationary distribution $\mu$, and ${\|h\|_{\mu} \le \kappa\|h\|_\rho}$ for every ${h : \mathcal{S} \to \mathbb{R}}$, for some $\kappa < \infty$.
\end{assumption}

The norm inequality is a coverage condition. It holds if and only if the density ratio $d\mu/d\rho$ is bounded by $\kappa^2$. If $\rho$ is stationary for $\Ppi$, that is, $\rho\Ppi = \rho$, Assumption~\ref{ass:stat-cov} holds with $\mu = \rho$ and $\kappa = 1$.

\begin{theorem}[Calibration-refinement bound]\label{thm:cal-ref}
Let $\vzeroGBC$ be the fixed point of $\TGBC$ from Lemma~\ref{lem:fixed-point}. Under Assumption~\ref{ass:stat-cov},
\[
  \|\vK - \vzero\|_{\mu}
  \;\le\; \|\vK - \vzeroGBC\|_{\mu}
  \;+\; \frac{\kappa}{1-\gamma}\,\|(I - \PiE)\,e\|_\rho.
\]
\end{theorem}

We call the first term the calibration error. It is the distance from $\vK$ to the fixed point of the coarsened Bellman operator. The second term is proportional to
\begin{equation}\label{eq:refinement-error}
  \|(I - \PiE)e\|_\rho \;=\; \min_{\theta}\,\|e - \theta \circ \ehat\|_\rho \;=\; \big(\E_\rho[\mathrm{Var}(e \mid \ehat)]\big)^{1/2},
\end{equation}
which we call the refinement error. The minimum is over all functions $\theta: \mathbb{R} \to \mathbb{R}$. The refinement error is the part of the prediction error $e$ that remains after the best correction of the form $\theta \circ \ehat$. Since $\|\vhat + \theta \circ \ehat - \vzero\|_\rho = \|e - \theta \circ \ehat\|_\rho$, it is also the smallest value of $\|v - \vzero\|_\rho$ over predictors $v$ of the form $\vhat + \theta \circ \ehat$.

BC's bound \citep[Theorem~1]{vanderlaan2025bc} has the same two terms with $\vhat$ as the conditioning axis, under BC's own stationary distribution. Its refinement error is the norm of $(I - \PiV)\vzero$, where $\PiV$ is the projection onto functions of $\vhat$. Since $\vzero = \vhat + e$ and $\PiV\vhat = \vhat$, we have $(I - \PiV)\vzero = (I - \PiV)e$, whose norm in $\|\cdot\|_\rho$ is $(\E_\rho[\mathrm{Var}(e \mid \vhat)])^{1/2}$. Comparing with Eq.~\eqref{eq:refinement-error}, the refinement error is governed by $\mathrm{Var}(e \mid \vhat)$ for BC and by $\mathrm{Var}(e \mid \ehat)$ for \GBC{}. The former is a constant determined by the base predictor, whereas the latter is small whenever $\ehat$ is close to $e$.

\subsection{Refinement Error Bounds via Error-Model Accuracy}\label{sec:refinement-training}

The second term in the calibration-refinement bound of Theorem~\ref{thm:cal-ref} is proportional to the refinement error $\|(I-\PiE)e\|_\rho$ of Eq.~\eqref{eq:refinement-error}.
We bound it in terms of the accuracy of $\ehat$.
What $\ehat$ estimates depends on its training target.
As described in Section~\ref{sec:gbc-algo}, $\ehat$ estimates $e$ itself under MC training and $(I-\gamma\Ppi)e$ under TD training.
We treat the two cases in turn.

\begin{proposition}[MC training]\label{prop:mc-refinement}
The refinement error satisfies
\[
  \|(I - \PiE)\,e\|_\rho \;\le\; \|\ehat - e\|_\rho.
\]
\end{proposition}

The bound follows from Eq.~\eqref{eq:refinement-error} by taking $\theta$ to be the identity.
Under MC training, the right-hand side is the root mean squared error of $\ehat$ under $\rho$, which may differ from the distribution of training states.

Under TD training, $\ehat$ estimates the Bellman residual $(I - \gamma\Ppi)e$ rather than the prediction error $e$.
Since $e = \sum_{t\ge0}\gamma^t\Ppi^t(I - \gamma\Ppi)e$, the prediction error at a state depends on the residuals at the states visited afterwards.
Even when $\ehat$ equals the residual exactly, $e$ need not be a function of $\ehat$, because $\Ppi$ need not map functions of $\ehat$ to functions of $\ehat$.
The following quantity measures how far it takes them outside this class.

\begin{definition}[Inherent Bellman error]\label{def:ibe-ehat}
Let $\FE := \{\theta \circ \ehat : \theta\colon \mathbb{R} \to \mathbb{R}\}$ be the class of functions of $\ehat$. We define
\[
  \BEhat := \sup_{f \in \FE,\, \|f\|_\infty \le 1}\;\inf_{g \in \FE}\; \|g - \Ppi f\|_\rho.
\]
\end{definition}

This is the inherent Bellman error \citep{munos2008fitted,chen2019information}
of $\Ppi$ on $\FE$. It lies in $[0, 1]$, since taking $g = 0$ gives
$\|\Ppi f\|_\rho \le \|f\|_\infty \le 1$. It is zero if and only if $\Ppi$ maps every bounded
function in $\FE$ to a function in $\FE$, that is, if and only if the distribution of
$\ehat(S')$ given $S$, with $S'$ the next state under $\pi$, depends on $S$ only through $\ehat(S)$.

When $\ehat$ does not equal the residual exactly, its error at the states visited afterwards matters, because $e$ depends on the residuals at those states.
This error and the inherent Bellman error are both measured under $\rho$, so $\rho$ has to cover these states.
We state this requirement using the discounted state-occupancy distribution of $\pi$ started from $\rho$, $d_{\pi,\rho} := (1-\gamma)\sum_{t\ge0}\gamma^t\rho\Ppi^t$.

\begin{assumption}[Occupancy coverage]\label{ass:occ-cov}
For some constant $\Cocc < \infty$, ${\|h\|_{d_{\pi,\rho}} \le \sqrt{\Cocc}\,\|h\|_\rho}$ for every ${h : \mathcal{S} \to \mathbb{R}}$.
\end{assumption}

Assumption~\ref{ass:occ-cov} bounds the density ratio of $d_{\pi,\rho}$ to $\rho$ by $\Cocc$.
If $\rho$ is stationary for $\Ppi$, then $d_{\pi,\rho} = \rho$ and Assumption~\ref{ass:occ-cov} holds with $\Cocc = 1$.

\begin{proposition}[TD training]\label{prop:td1-refinement}
Suppose $\ehat$ is bounded and Assumption~\ref{ass:occ-cov} holds. Then
\[
  \|(I - \PiE)\,e\|_\rho
  \;\le\; \sqrt{\Cocc}\!\left[\frac{\|\ehat - (I - \gamma\Ppi)e\|_\rho}{1-\gamma}
  + \frac{\gamma\,\BEhat\,\|\ehat\|_\infty}{(1-\gamma)^2}\right].
\]
\end{proposition}

The second term does not vanish as $\ehat$ becomes more accurate, so the bound is informative only when $\BEhat$ is small.\footnote{Appendix~\ref{app:next-bin-diagnostic} reports a diagnostic for $\BEhat$ in our experiments.}
When $\BEhat = 0$, the bound is $\sqrt{\Cocc}/(1-\gamma)$ times the error of $\ehat$ in estimating the residual, whereas Proposition~\ref{prop:mc-refinement} bounds the refinement error by the error of $\ehat$ in estimating $e$.

\subsection{Refinement Error Bounds under Low-Dimensional Structure}\label{sec:lowdim-adaptation}

We now specialize the bounds of Section~\ref{sec:refinement-training} to the case where the function that $\ehat$ estimates has low-dimensional structure.
We write $f^*$ for this function, that is, $f^* = e$ under MC training and
$f^* = (I-\gamma\Ppi)e$ under TD training.

\begin{assumption}[Approximate low-dimensional structure]\label{ass:LS1}
There exist a map $\phi : \mathcal{S} \to \mathbb{R}^k$, a $\beta$-H\"older
function $g : \mathbb{R}^k \to \mathbb{R}$ for some smoothness exponent
$\beta > 0$, and a constant $\estar \ge 0$ such that
$\|f^* - g\circ\phi\|_\rho \le \estar$.
\end{assumption}

\begin{assumption}[Convergence rate of the error model]\label{ass:LS2}
For $g$, $\phi$, $\beta$, and $k$ as in Assumption~\ref{ass:LS1}, the error model $\ehat$ satisfies
\[
  \|\ehat - g\circ\phi\|_\rho^2
  \;\le\; \Cadapt\,\non^{-2\beta/(2\beta+k)}
  \;+\; \siglab^2/\non
\]
with probability at least $1 - \delta$, where $\Cadapt < \infty$ is a constant and $\siglab^2$ bounds the conditional
variance of the training target given $S$.
\end{assumption}

Assumption~\ref{ass:LS1} states that $f^*$, the function that $\ehat$ estimates, is approximately
a H\"older function of a $k$-dimensional summary $\phi(S)$ of the state, with approximation error at most $\estar$.
We call $k$ the intrinsic dimension.
When $\mathcal{S} \subseteq \mathbb{R}^d$, the case of interest is $k$ much smaller than the state dimension $d$.
Assumption~\ref{ass:LS2} states that the squared error of $\ehat$ in estimating $g\circ\phi$ decreases at the rate $\non^{-2\beta/(2\beta+k)}$, which does not depend on $d$.
This is the minimax-optimal rate for $\beta$-H\"older regression in $k$ dimensions
\citep{stone1980optimal,stone1982optimal}.
Appendix~\ref{app:rw-lowdim} lists estimators for which rates of this form have been established.

\begin{theorem}[Refinement bound under low-dimensional structure, MC]\label{thm:rate-transfer}
Suppose $\ehat$ is obtained by MC training and
Assumptions~\ref{ass:LS1} and~\ref{ass:LS2} hold with $f^* = e$. Then with
probability at least $1 - \delta$,
\[
  \|(I - \PiE)\,e\|_\rho
  \;\le\; \sqrt{\Cadapt}\,\non^{-\beta/(2\beta+k)}
  \;+\; \frac{\siglab}{\sqrt{\non}}
  \;+\; \estar.
\]
\end{theorem}

The bound follows from Proposition~\ref{prop:mc-refinement} and the triangle inequality
$\|\ehat - e\|_\rho \le \|\ehat - g\circ\phi\|_\rho + \|g\circ\phi - e\|_\rho$, where
Assumption~\ref{ass:LS2} bounds the first term and Assumption~\ref{ass:LS1} bounds the second.
The proof is in Appendix~\ref{app:proof-thm4}.
The refinement error therefore inherits the rate of the error model, which depends on the intrinsic dimension $k$ and not on the state dimension $d$.
Whether Assumptions~\ref{ass:LS1} and~\ref{ass:LS2} hold depends on the problem and on the estimator used to obtain $\ehat$.
If either fails, the rate in Theorem~\ref{thm:rate-transfer} is not guaranteed.

\begin{remark}[BC's refinement error under Assumption~\ref{ass:LS1}]\label{rem:bc-failure}
Even when Assumption~\ref{ass:LS1} holds with $f^* = e$, BC's refinement error need not be small.
As shown in Section~\ref{sec:cal-ref-decomp}, this error is $(\E_\rho[\mathrm{Var}(e \mid \vhat)])^{1/2}$, a constant determined by the base predictor.
For example, suppose $\vzero = \vhat + g\circ\phi$, so that $e = g\circ\phi$ satisfies Assumption~\ref{ass:LS1} with $\estar = 0$.
If $\vhat(S)$ and $\phi(S)$ are independent under $\rho$, then
$\E_\rho[\mathrm{Var}(e \mid \vhat)] = \mathrm{Var}_\rho(e)$, so conditioning on $\vhat$ removes none of the variance of $e$.
In contrast, if Assumption~\ref{ass:LS2} holds in this example, Theorem~\ref{thm:rate-transfer} shows that \GBC{}'s refinement error goes to zero as $\non$ grows.
\end{remark}

We now return to TD training.
Repeating the proof of Theorem~\ref{thm:rate-transfer} with Proposition~\ref{prop:td1-refinement} in place of Proposition~\ref{prop:mc-refinement} gives the following bound.

\begin{theorem}[Refinement bound under low-dimensional structure, TD]\label{cor:rate-transfer-td1}
Suppose $\ehat$ is obtained by TD training and
Assumptions~\ref{ass:LS1} and~\ref{ass:LS2} hold with $f^* = (I - \gamma\Ppi)e$.
Suppose also that $\ehat$ is bounded and Assumption~\ref{ass:occ-cov} holds, as in Proposition~\ref{prop:td1-refinement}.
Then with probability at least $1 - \delta$,
\[
  \|(I - \PiE)\,e\|_\rho
  \;\le\; \sqrt{\Cocc}\!\left[\frac{\sqrt{\Cadapt}\,\non^{-\beta/(2\beta+k)}
  + \siglab/\sqrt{\non} + \estar}{1-\gamma}
  + \frac{\gamma\,\BEhat\,\|\ehat\|_\infty}{(1-\gamma)^2}\right].
\]
\end{theorem}

The proof is in Appendix~\ref{app:cor41}.
Inside the brackets, the first term is the bound of Theorem~\ref{thm:rate-transfer} divided by $1-\gamma$, and the second term is the same as in Proposition~\ref{prop:td1-refinement}.
Assumptions~\ref{ass:LS1} and~\ref{ass:LS2} do not imply that the inherent Bellman error $\BEhat$ is small, because $\BEhat$ depends on how $\Ppi$ acts on functions of $\ehat$.

\subsection{Finite-Sample Bounds}\label{sec:finite-sample}

We now account for the $B$ bins and the finite calibration sample used by Algorithm~\ref{alg:gbc}.
We give two results.
Theorem~\ref{thm:est-error} bounds the estimation error for a given error model $\ehat$.
Theorem~\ref{thm:combined} combines it with the refinement bounds of Section~\ref{sec:lowdim-adaptation}, which assume low-dimensional structure in the function that $\ehat$ estimates.

Let $\PiEB$ denote the $L^2(\rho)$ projection onto the histogram class $\FBE := \{\theta \circ \ehat : \theta \in \FB\}$, that is, ${(\PiEB q)(s) = \E_\rho[q(S) \mid \ehat(S) \in I_j]}$ for $\ehat(s) \in I_j$.
Replacing $\PiE$ by $\PiEB$ in Eq.~\eqref{eq:coarsened-op} gives the binned version of the coarsened Bellman operator, $\TGBCB(g) := \vhat + \PiEB[\Tpi(g) - \vhat]$, which has a unique bounded fixed point $\vzeroGBCB$ by the argument of Lemma~\ref{lem:fixed-point}.
Algorithm~\ref{alg:gbc} is an empirical version of the iteration $\hat{v}^{(k+1)} = \TGBCB(\hat{v}^{(k)})$, with the true Bellman operator $\Tpi$ replaced by the DR target $\Tpihat$ and $\PiEB$ by binwise sample means.

The finite-sample results use four conditions.
Conditions C1--C3 in Appendix~\ref{app:thm2} require bounded rewards, weights, $\vhat$, $\ehat$, and iterates $\hat{v}^{(k)}$ (C1), i.i.d.\ transitions in $\Doff$ that are independent of the data used to obtain $\vhat$, $\ehat$, $\wpihat$, $\hat{r}$, and $\hat{P}$ (C2), and a bounded linear operator $\hat{P}$ estimating the transition kernel $P$ (C3).
Condition C4 in Appendix~\ref{app:lemmas-78} is Assumption~\ref{ass:stat-cov} with the coarsened transition operator $\PpiE$ of Eq.~\eqref{eq:coarsened-trans} replaced by its binned version $\PpiEB := \PiEB\Ppi$.
That is, $\PpiEB$ has a stationary distribution $\mu_B$, and ${\|h\|_{\mu_B} \le \kappa_B\|h\|_\rho}$ for every ${h : \mathcal{S} \to \mathbb{R}}$, for some $\kappa_B < \infty$.

\begin{theorem}[Estimation error bound]\label{thm:est-error}
Under conditions C1--C4 (Appendices~\ref{app:thm2} and~\ref{app:lemmas-78}),
there is a constant $C \in (0, \infty)$ such that, for any number of iterations $K$ and
with probability at least $1 - \delta$,
\begin{align*}
  \|\vK - \vzero\|_{\mu_B}
  &\le \underbrace{\frac{\kappa_B}{1-\gamma}\,\|(I - \PiEB)\,e\|_\rho}_{\termRB\ \text{binned refinement}}
    + \underbrace{\frac{C\kappa_B}{1-\gamma}\left[\sqrt{\frac{B}{\noff}\log\frac{\noff}{B}}
    + \sqrt{\frac{\log(K/\delta)}{\noff}}\right]}_{\termS\ \text{statistical}}\\
  &\quad + \underbrace{\frac{\kappa_B}{1-\gamma}\,\max_{0 \le k < K}
      \big\|(\wpihat - \wpi)(\hat q_{\hat v^{(k)}} - q_{\hat v^{(k)}})\big\|_{\rho,b}}_{\termN\ \text{DR nuisance product}}
    + \underbrace{\gamma^K\,\|\hat{v}^{(0)} - \vzeroGBCB\|_{\mu_B}}_{\termI\ \text{iteration}},
\end{align*}
where $q_v := r + \gamma P v$, $\hat q_v := \hat{r} + \gamma \hat{P} v$, and $\|\cdot\|_{\rho,b}$ is the $L^2$ norm under $S \sim \rho$ and $A \sim b(\cdot \mid S)$.
\end{theorem}

The proof is in Appendix~\ref{app:lemmas-78}.
The bound has the form of Theorem~\ref{thm:cal-ref}, with $\PiE$, $\kappa$, and $\mu$ replaced by their binned versions.
The term $\termRB$ is proportional to the binned refinement error $\|(I - \PiEB)e\|_\rho$, the part of the prediction error $e$ that remains after the best correction that is constant on each bin.
This error depends on the accuracy of $\ehat$ and does not decrease with $\noff$ or $K$.
The other three terms bound the calibration error $\|\vK - \vzeroGBCB\|_{\mu_B}$, the distance from $\vK$ to the fixed point of the binned operator.
Of these, $\termS$ is the statistical error of the histogram regression on the $\noff$ calibration transitions, $\termN$ contains the product of the errors of the importance weights $\wpihat$ and the estimate $\hat q_v$ of $q_v$, and $\termI$ is $\gamma^K$ times the distance from the initial iterate $\hat{v}^{(0)} := \vhat$ to this fixed point.
Theorem~\ref{thm:est-error} is the counterpart of BC's bound \citep[Theorem~4]{vanderlaan2025bc}, with $\ehat$ in place of $\vhat$ as the conditioning axis.
As in that result, it does not assume Bellman completeness, because $\TGBCB$ maps the class $\vhat + \FBE$ to itself by construction.

In Theorem~\ref{thm:est-error}, the error model $\ehat$ is given, and its accuracy enters through the binned refinement error.
The next theorem bounds this error in terms of $\non$ when $\ehat$ is trained on the on-policy sample under the low-dimensional structure of Section~\ref{sec:lowdim-adaptation}.

\begin{theorem}[Estimation error bound under low-dimensional structure]\label{thm:combined}
Suppose the assumptions of Theorem~\ref{thm:est-error} hold.
Suppose also that the function $x \mapsto \E_\rho[e \mid \ehat(S) = x]$ has a version of total variation at most $\CTV$ on the range of $\ehat$.
Suppose further that ${\rho(\ehat(S) \in I_j) \le c_{\mathrm{bin}}/B}$ for every $j$.
The constants $\CTV$ and $c_{\mathrm{bin}}$ do not depend on $B$, $K$, $\non$, or $\noff$.
Then, in each of the two cases below, for any $K \in \mathbb{N}$ and with probability at least $1 - 2\delta$,
\[
  \|\vK - \vzero\|_{\mu_B} \;\le\; \termR + \termW + \termS + \termN + \termI,
  \qquad
  \termW := \frac{\kappa_B\,\CTV}{2(1-\gamma)}\sqrt{\frac{c_{\mathrm{bin}}}{B}},
\]
where $\termS$, $\termN$, and $\termI$ are the terms of Theorem~\ref{thm:est-error} and $\termR$ is given in each case.

\textup{(i)} \emph{MC training.} Under the assumptions of Theorem~\ref{thm:rate-transfer},
\[
  \termR := \termRMC :=\frac{\kappa_B}{1-\gamma}\left[\sqrt{\Cadapt}\,
      \non^{-\beta/(2\beta+k)} + \frac{\siglab}{\sqrt{\non}} + \estar\right].
\]
\textup{(ii)} \emph{TD training.} Under the assumptions of Theorem~\ref{cor:rate-transfer-td1},
\[
  \termR := \termRTD :=\sqrt{\Cocc}\left[\frac{\termRMC}{1-\gamma}
  + \frac{\kappa_B\,\gamma\,\BEhat\,\|\ehat\|_\infty}{(1-\gamma)^3}\right].
\]
\end{theorem}

The proofs of cases (i) and (ii) are in Appendices~\ref{app:proof-thm5} and~\ref{app:cor51}, respectively.
Case (ii) covers the TD training used in our experiments.
The bound is that of Theorem~\ref{thm:est-error} with $\termRB$ replaced by $\termR + \termW$.
Here $\termR$ is proportional to the bound on the refinement error given by Theorem~\ref{thm:rate-transfer} in case (i) and by Theorem~\ref{cor:rate-transfer-td1} in case (ii), and $\termW$ accounts for the increase in the refinement error caused by the bins.
The term $\termW$ is proportional to $\CTV\sqrt{c_{\mathrm{bin}}/B}$, which decreases at the rate $B^{-1/2}$ as the number of bins grows.
The constant $\CTV$ bounds the total variation of $x \mapsto \E_\rho[e \mid \ehat(S) = x]$, which is the minimizer $\theta$ in Eq.~\eqref{eq:refinement-error}.
The total variation of a function $m$ is the total amount by which it rises and falls, that is, the supremum of $\sum_{i=1}^{n} |m(x_i) - m(x_{i-1})|$ over all finite sequences ${x_0 < \dots < x_n}$ in its domain.
The assumption on the bins holds with $c_{\mathrm{bin}} = 1$ when the bins have equal probability under $\rho$, as with bin edges at the quantiles of $\ehat(S)$.

To obtain a rate from Theorem~\ref{thm:combined}, we choose $K$ and $B$.
Suppose that $\kappa_B$ is bounded uniformly in $B$, and let $K \asymp \log\noff$ with a proportionality constant large enough that the term $\termI$ is at most $\termS$.
Then $\termW = O(B^{-1/2})$ and ${\termS = O\big(((B/\noff)\log(\noff/B))^{1/2}\big)}$, so a larger $B$ reduces $\termW$ and increases $\termS$.
Taking $B \asymp (\noff/\log\noff)^{1/2}$ makes both bounds of order $(\log\noff/\noff)^{1/4}$.
If $\termN$ is at most of this order and $\estar = 0$, then Theorem~\ref{thm:combined} gives
\begin{align*}
  \text{MC training:}\quad \|\vK - \vzero\|_{\mu_B}
  &= O\!\left(\frac{1}{1-\gamma}\left[
    \non^{-\beta/(2\beta+k)} + \Big(\frac{\log\noff}{\noff}\Big)^{1/4}
  \right]\right),\\
  \text{TD training:}\quad \|\vK - \vzero\|_{\mu_B}
  &= O\!\left(\frac{1}{1-\gamma}\left[
    \frac{\non^{-\beta/(2\beta+k)}}{1-\gamma} + \frac{\BEhat}{(1-\gamma)^2}
    + \Big(\frac{\log\noff}{\noff}\Big)^{1/4}
  \right]\right).
\end{align*}
The MC bound goes to zero as $\non$ and $\noff$ grow.
The TD bound also contains the inherent Bellman error $\BEhat$ of Definition~\ref{def:ibe-ehat}, which need not decrease with either sample size.

The two sample sizes enter the bound of Theorem~\ref{thm:combined} through different terms.
The on-policy sample size $\non$ and the low-dimensional structure appear only in the refinement term $\termR$, and the calibration sample size $\noff$ appears only in the statistical term $\termS$.
In BC \citep{vanderlaan2025bc}, by contrast, the refinement error is a constant determined by the base predictor (Remark~\ref{rem:bc-failure}) and does not decrease with either sample size.

\section{Experiments}\label{sec:experiments}

We evaluate \GBC{} on three environments: the CRM benchmark of \citet{vanderlaan2025bc}, a controlled synthetic MDP (Toy), and three D4RL continuous-control tasks \citep{fu2020d4rl} (HalfCheetah, Hopper, Walker2d). We compare Raw (the uncalibrated base predictor $\vhat$), BC, Direct (the predictor $\vhat+\ehat$), and \GBC{}, reporting RMSE against a Monte-Carlo ground truth $\vzero$ under $\pi$. In every environment, BC receives the same on-policy sample that \GBC{} uses to fit $\ehat$, appended to its calibration data with importance weight one, so that both methods have access to the same data. The on-policy sample $\Don$ is smaller than the off-policy calibration sample $\Doff$: $\non = 100$ against $\noff = 5{,}000$ transitions on Toy, about $10^4$ against $10^5$ on CRM, and about $10^3$ against $10^4$ on D4RL. The defaults are $B = 32$ bins, $K = 3$ calibration iterations, and TD training for $\ehat$ (Section~\ref{sec:gbc-algo}). The base predictor is fit on off-policy data disjoint from $\Doff$. Further details are in Appendix~\ref{app:impl}. Within each seed, all methods share the base predictor and the states on which RMSE is computed, so we also use paired differences, the mean $\pm$ standard error (SE) over seeds of the per-seed differences in RMSE between two methods. In the tables below, bold marks the method with the lowest mean RMSE and any method whose paired difference from it is within two SE. A KuaiRec-based semi-synthetic recommender evaluation is in Appendix~\ref{app:kuairec}.

\paragraph{CRM.}
We inherit the CRM environment of \citet{vanderlaan2025bc}, with a six-dimensional customer state, three promotion actions, and $\gamma = 0.99$. The policies are adapted from theirs. The target policy is an aggressive deterministic promotion rule, and the behavior policy mixes it with a heuristic promotion rule (Appendix~\ref{app:impl}). We evaluate three base predictors (linear ridge, boosted trees, neural MLP) at different numbers of fitting iterations. The base predictor $\vhat$ is fit by fitted value iteration \citep{munos2008fitted} with all importance weights set to one, so it estimates the value of the behavior policy rather than $\vzero$. Appendix~\ref{app:additional-exp-details} reports the results when $\vhat$ is fit with importance weights. \GBC{} has the lowest mean RMSE among the methods in Table~\ref{tab:crm-headline} in all four settings, improving over BC by $4$--$15\%$ and over Direct by $9$--$20\%$. Even for the linear predictor with $18$ fitting iterations, which has the lowest Raw RMSE ($0.360$), \GBC{} lowers the RMSE to $0.298$, whereas BC reaches $0.324$. We also measure, on the same runs, how well each conditioning axis groups states with similar prediction errors (Appendix~\ref{app:origin-diagnostic}). For the neural predictor, where \GBC{} improves most over BC, binning by $\ehat$ reduces the within-bin standard deviation of the true prediction error $e$ by $64\%$, against $36\%$ for binning by $\vhat$. Since a binwise correction leaves this within-bin variation, this is consistent with the refinement error of Eq.~\eqref{eq:refinement-error} being smaller with $\ehat$ than with $\vhat$ as the conditioning axis.

\begin{table}[h]
\centering
\caption{CRM RMSE scaled by $1 - \gamma$ (mean $\pm$ SE over $10$ seeds).}
\label{tab:crm-headline}
\small
\begin{tabular}{llcccc}
\toprule
Base $\vhat$ & $\vhat$ iter. & Raw & BC & Direct & \GBC{} (ours) \\
\midrule
Linear  &  $8$ & $0.718 \pm 0.009$ & $0.520 \pm 0.008$ & $0.627 \pm 0.010$ & $\mathbf{0.500 \pm 0.007}$ \\
Linear  & $18$ & $0.360 \pm 0.007$ & $0.324 \pm 0.007$ & $0.325 \pm 0.008$ & $\mathbf{0.298 \pm 0.006}$ \\
Boosted & $18$ & $0.807 \pm 0.011$ & $0.631 \pm 0.010$ & $0.705 \pm 0.011$ & $\mathbf{0.566 \pm 0.009}$ \\
Neural  & $100$ & $0.833 \pm 0.016$ & $0.718 \pm 0.023$ & $0.741 \pm 0.014$ & $\mathbf{0.608 \pm 0.011}$ \\
\bottomrule
\end{tabular}
\end{table}

\paragraph{Toy MDP.}
Motivated by the low-dimensional structure of Section~\ref{sec:lowdim-adaptation}, we design the Toy MDP so that the value depends on only two of the $d = 10$ state coordinates. The reward (with a sine in $x_2$), both policies, and the dynamics of $x_1, x_2$ depend only on $(x_1, x_2)$, while the other coordinates $x_3,\dots,x_d$ are noise, redrawn independently at every step. We set $\gamma = 0.8$. We compare calibration along $\vhat$ and $\ehat$ for a linear base predictor, which is fit as on CRM and cannot represent the nonlinear value function. Since $x_3,\dots,x_d$ are noise, the coefficients of $\vhat$ on them are zero in the population limit. The function that $\ehat$ estimates then depends only on $(x_1, x_2)$, and approximately so with finite data (Assumption~\ref{ass:LS1} with intrinsic dimension $k = 2$). \GBC{} improves on Raw, whereas BC and Direct do not (Table~\ref{tab:toy-headline}).

\begin{table}[h]
\centering
\caption{Toy MDP RMSE (mean $\pm$ SE over $10$ seeds).}
\label{tab:toy-headline}
\small
\begin{tabular}{lc}
\toprule
Method & RMSE \\
\midrule
Raw                        & $0.429 \pm 0.005$ \\
BC                         & $0.451 \pm 0.006$ \\
Direct                     & $0.474 \pm 0.036$ \\
\GBC{} (ours)            & $\mathbf{0.364 \pm 0.010}$ \\
\bottomrule
\end{tabular}
\end{table}

\paragraph{D4RL.}
To test \GBC{} on standard continuous-control tasks, we use the \texttt{simple-v0}, \texttt{medium-v0}, and \texttt{expert-v0} MuJoCo datasets from Minari, the successor to D4RL for offline RL datasets \citep{minari}. The base predictor is a Monte-Carlo regression of returns on \texttt{simple-v0} trajectories. The calibration sample comes from \texttt{medium-v0} (behavior policy), and the on-policy sample and held-out evaluation returns from \texttt{expert-v0} (target policy), with $\gamma = 0.9$. Table~\ref{tab:d4rl-headline} reports per-task RMSE. \GBC{} and Direct improve over Raw on all three tasks by more than three SE in paired differences. BC has a higher mean RMSE than Raw on all three tasks. \GBC{} does not uniformly improve on Direct. The paired differences (\GBC{} minus Direct) show that \GBC{} is similar to Direct on HalfCheetah ($-0.45 \pm 0.36$) and Hopper ($-0.02 \pm 0.02$), and more accurate on Walker2d ($-0.10 \pm 0.03$).
\begin{table}[h]
\centering
\caption{D4RL RMSE (mean $\pm$ SE over $10$ seeds).}
\label{tab:d4rl-headline}
\small
\begin{tabular}{lccc}
\toprule
Method & HalfCheetah & Hopper & Walker2d \\
\midrule
Raw                   & $25.66 \pm 0.81$ & $4.95 \pm 0.17$ & $7.20 \pm 0.23$ \\
BC                    & $29.48 \pm 1.10$ & $5.02 \pm 0.17$ & $7.44 \pm 0.23$ \\
Direct         & $\mathbf{23.43 \pm 0.75}$ & $\mathbf{4.60 \pm 0.15}$ & $6.79 \pm 0.22$ \\
\GBC{} (ours)       & $\mathbf{22.98 \pm 0.68}$ & $\mathbf{4.58 \pm 0.15}$ & $\mathbf{6.69 \pm 0.22}$ \\
\bottomrule
\end{tabular}
\end{table}

\paragraph{Ablations.}
Appendix~\ref{app:additional-exp-details} varies $K$, $B$, and $\non$ on Toy one at a time from the defaults. \GBC{} has the lowest RMSE at every value tried. Its RMSE decreases with $K$ until $K = 10$, consistent with the iteration term $\termI$ of Theorem~\ref{thm:est-error}, which decays as $\gamma^K$. BC becomes more accurate than Raw from $K = 10$, and Direct at $\non = 400$. On D4RL, we also fit $\ehat$ by MC training, with returns truncated at the end of each on-policy window. The RMSE of \GBC{} changes by less than $2\%$, whereas Direct becomes the most accurate method on HalfCheetah but less accurate than Raw on Hopper and Walker2d. This is consistent with \GBC{} using $\ehat$ only to group states, whereas Direct adds $\ehat$ to $\vhat$.

\section{Conclusion}\label{sec:conclusion}

We study semi-offline value estimation, where a small on-policy sample supplements a large off-policy dataset and the goal is to correct a given value predictor $\vhat$. We propose Error-Guided Bellman Calibration (GBC), which calibrates $\vhat$ along a learned error axis $\ehat$ via an additive update $\vhat + \theta \circ \ehat$. \GBC{} trains $\ehat$, a model of the prediction error of $\vhat$, on the on-policy sample and fits the correction $\theta$ on the off-policy dataset. We bound the refinement term, which measures the part of the prediction error that calibration along $\ehat$ cannot remove, in terms of the accuracy of $\ehat$. Under an assumed error-model rate and low-dimensional structure in the function that $\ehat$ estimates, this bound adapts to the intrinsic dimension rather than the state dimension. Combining it with a finite-sample analysis of the calibration step, we bound the estimation error of \GBC{}. In all three experimental environments, \GBC{} reduces the error of the given predictor and improves on the baselines in most settings.

\paragraph{Limitations.}
We do not establish that the low-dimensional structure of Assumption~\ref{ass:LS1} is generally present. Our rates in $\non$ require this structure and an assumed convergence rate of $\ehat$ (Assumption~\ref{ass:LS2}). With TD training, used in our experiments, the refinement bound of Proposition~\ref{prop:td1-refinement} carries an inherent Bellman error term and needs a state-coverage condition. The finite-sample results of Theorem~\ref{thm:est-error} assume independence, boundedness, and coverage conditions and a fixed partition, which our implementation does not ensure (Appendix~\ref{app:impl}). In our experiments, \GBC{} does not uniformly improve on Direct. We do not analyze training $\ehat$ on off-policy data, which is possible in principle with importance weights.

\begin{ack}
The author received no external funding for this work and declares no competing interests.
\end{ack}

\bibliographystyle{plainnat}
\bibliography{refs}

\appendix

\section{Proofs}\label{app:proofs}

Throughout the appendices, $\|\cdot\|$ without a subscript denotes $\|\cdot\|_\rho$.

\subsection{Proof of Lemma~\ref{lem:fixed-point}}\label{app:proof-lem1}

The argument is the standard one for a supremum-norm non-expansion composed with a Bellman operator \citep{munos2008fitted}.
We take $\PiE$ to be the Markov operator given by a regular conditional distribution of $S$ given $\ehat(S)$ under $\rho$. This fixes a version of the conditional expectation that defines $\PiE$ in Section~\ref{sec:cal-ref-decomp}. For every bounded measurable function $q$ on $\mathcal{S}$, $\PiE q$ is then a function of $\ehat$ and $\|\PiE q\|_\infty \le \|q\|_\infty$. The Markov kernel $\Ppi$ satisfies $\|\Ppi q\|_\infty \le \|q\|_\infty$ as well.

Because $r_\pi$ and $\vhat$ are bounded, $\TGBC$ maps bounded measurable functions to bounded measurable functions. For two such functions $g_1$ and $g_2$, the linearity of $\PiE$ and the identity $\Tpi(g_1) - \Tpi(g_2) = \gamma\Ppi(g_1 - g_2)$ give
\[
  \TGBC(g_1) - \TGBC(g_2) = \gamma\,\PiE\Ppi(g_1 - g_2),
\]
and therefore $\|\TGBC(g_1) - \TGBC(g_2)\|_\infty \le \gamma\,\|g_1 - g_2\|_\infty$. Thus $\TGBC$ is a $\gamma$-contraction on the space of bounded measurable functions, which is complete under $\|\cdot\|_\infty$. The Banach fixed-point theorem then gives a unique bounded fixed point $\vzeroGBC$.

For the last claim, the fixed-point equation $\vzeroGBC = \TGBC(\vzeroGBC)$ reads $\vzeroGBC - \vhat = \PiE[\Tpi(\vzeroGBC) - \vhat]$, and the right-hand side is a function of $\ehat$. \qed

\subsection{Proof of Theorem~\ref{thm:cal-ref}}\label{app:proof-thm1}

The fixed-point characterization $\vzeroGBC - \hat{v} = \PiE[\Tpi(\vzeroGBC) -
\hat{v}]$ together with $\vzero = \Tpi \vzero$ gives, using
$\Tpi(g) - \Tpi(\vzero) = \gamma\Ppi(g - \vzero)$ and $\PpiE = \PiE \Ppi$,
\begin{align*}
  \vzeroGBC - \vzero
  &= \PiE[e + \gamma\Ppi(\vzeroGBC - \vzero)] - e\\
  &= -(I - \PiE)\,e + \gamma\,\PpiE(\vzeroGBC - \vzero).
\end{align*}
Rearranging,
\begin{equation}
  (I - \gamma\PpiE)\,(\vzeroGBC - \vzero) \;=\; -(I - \PiE)\,e.
  \label{eq:thm1-fp}
\end{equation}
Stationarity of $\mu$ and Jensen's inequality make $\PpiE$
non-expansive in $L^2(\mu)$.
Applying this norm to
Eq.~\eqref{eq:thm1-fp} and using triangle inequality with non-expansion,
\[
  \|\vzeroGBC - \vzero\|_{\mu}
  \;\le\; \gamma\,\|\vzeroGBC - \vzero\|_{\mu}
  + \|(I - \PiE)\,e\|_{\mu},
\]
so
\begin{equation}
  \|\vzeroGBC - \vzero\|_{\mu}
  \le \frac{\|(I - \PiE)e\|_{\mu}}{1-\gamma}
  \le \frac{\kappa}{1-\gamma}\,\|(I - \PiE)e\|.
  \label{eq:thm1-fp-bound}
\end{equation}
The triangle inequality in $L^2(\mu)$ yields the claim. \qed

\paragraph{Stationary special case.}
Conditional expectation gives $\rho\PiE=\rho$ and $\rho\PiEB=\rho$.
Thus $\rho(\PiE\Ppi)=\rho\Ppi$ and likewise for the bin operator: $\rho$ is stationary for either coarsened operator exactly when $\rho\Ppi=\rho$.
In that case both coverage coefficients may be taken as one.

\subsection{Proof of Proposition~\ref{prop:mc-refinement}}\label{app:proof-prop11}

The projection $\PiE$ is the $L^2(\rho)$-orthogonal projection onto the closed
subspace $\{\theta\circ\ehat : \theta \text{ measurable}\}$. The identity
calibrator $\theta_*(x) = x$ satisfies $\theta_*\circ\ehat = \ehat$, so $\ehat$
lies in the projection range. Projection optimality yields
$\|(I - \PiE)\,e\| = \min_\theta \|e - \theta\circ\ehat\| \le \|e - \ehat\|$. \qed

\subsection{Proof of Proposition~\ref{prop:td1-refinement}}\label{app:proof-prop12}

Set $\eta_{\mathrm{TD}} := (I - \gamma\Ppi)e - \ehat$, so that $(I - \gamma\Ppi)e
= \ehat + \eta_{\mathrm{TD}}$.

\emph{Step 1 (Discounted occupancy and the inverse operator).}
On bounded functions, let $\mathcal R_\pi:=(I-\gamma\Ppi)^{-1}=\sum_{t\ge0}\gamma^t\Ppi^t$; the series converges in the sup norm.
The operator $(1-\gamma)\mathcal R_\pi$ is a Markov kernel and takes the initial distribution $\rho$ to $d_{\pi,\rho}$.
Jensen's inequality and Assumption~\ref{ass:occ-cov} give, for every bounded $h$,
\begin{equation}\label{eq:prop12-resolvent}
  \|\mathcal R_\pi h\|^2
  \le \frac{\int h^2\,d d_{\pi,\rho}}{(1-\gamma)^2}
  \le \frac{\Cocc}{(1-\gamma)^2}\,\|h\|^2.
\end{equation}

\emph{Step 2 (A comparison function of $\ehat$).}
Recall $\PpiE=\PiE\Ppi$ and set $u:=(I-\gamma\PpiE)^{-1}\ehat$.
Since $\PpiE$ is a Markov operator, its geometric series converges in the sup norm, $u$ is a function of $\ehat$, and $\|u\|_\infty\le\|\ehat\|_\infty/(1-\gamma)$.
Projection optimality under $\rho$ and Definition~\ref{def:ibe-ehat} imply
\begin{equation}\label{eq:prop12-ibe}
  \|(\Ppi-\PpiE)u\|
  =\|(I-\PiE)\Ppi u\|
  \le\BEhat\|u\|_\infty
  \le\frac{\BEhat\|\ehat\|_\infty}{1-\gamma}.
\end{equation}
The inequality is immediate if $u=0$; otherwise apply the definition to $u/\|u\|_\infty$.

\emph{Step 3 (Refinement).}
Using $(I-\gamma\PpiE)u=\ehat$ and the definition of $\eta_{\mathrm{TD}}$ gives
\[
  (I-\gamma\Ppi)(e-u)
  =\eta_{\mathrm{TD}}+\gamma(\Ppi-\PpiE)u.
\]
Since $e$, $\ehat$, and $u$ are bounded, applying $\mathcal R_\pi$ to both sides gives
\[
  e-u
  =\mathcal R_\pi\big[\eta_{\mathrm{TD}}+\gamma(\Ppi-\PpiE)u\big].
\]
Because $u$ lies in the range of $\PiE$, projection optimality and
Eqs.~\eqref{eq:prop12-resolvent}--\eqref{eq:prop12-ibe} yield
\[
  \|(I-\PiE)e\|\le\|e-u\|
  \le\sqrt{\Cocc}\left[
    \frac{\|\eta_{\mathrm{TD}}\|}{1-\gamma}
    +\frac{\gamma\BEhat\|\ehat\|_\infty}{(1-\gamma)^2}\right].
\]
Since $\|\eta_{\mathrm{TD}}\| = \|\ehat - (I - \gamma\Ppi)e\|$, this is the claim. \qed

\paragraph{Interpretation of Assumption~\ref{ass:occ-cov}.}
For a finite state space, Assumption~\ref{ass:occ-cov} is the bound
$\max_{s:\rho(s)>0}d_{\pi,\rho}(s)/\rho(s)\le\Cocc$, together with zero occupancy wherever $\rho(s)=0$.
If $\rho\Ppi=\rho$, then $d_{\pi,\rho}=\rho$ and $\Cocc=1$ is admissible.
Finite coverage also allows unequal distributions; for example, a two-state chain moving to and remaining at state 2, started from the uniform distribution, has occupancy $((1-\gamma)/2,(1+\gamma)/2)$ and admits $\Cocc=1+\gamma$.
Bounded action weights alone do not imply this condition, since even a single-action chain can visit states absent from the initial logging distribution.

\subsection{Proof of Theorem~\ref{thm:rate-transfer}}\label{app:proof-thm4}

By the triangle inequality in $L^2(\rho)$,
\begin{equation}
  \|\ehat - e\|
  \;\le\; \|\ehat - g\circ\phi\| + \|g\circ\phi - e\|.
  \label{eq:thm4-triangle}
\end{equation}
Assumption~\ref{ass:LS1} (with $f^* = e$) gives
$\|g\circ\phi - e\| \le \estar$.
Assumption~\ref{ass:LS2}, on its high-probability event, gives
$\|\ehat - g\circ\phi\|^2 \le \Cadapt\non^{-2\beta/(2\beta+k)}
+ \siglab^2/\non$. Using $\sqrt{a+b} \le \sqrt{a} + \sqrt{b}$ for $a, b \ge 0$,
\begin{equation}
  \|\ehat - g\circ\phi\|
  \;\le\; \sqrt{\Cadapt}\,\non^{-\beta/(2\beta+k)}
  + \siglab/\sqrt{\non}.
  \label{eq:thm4-LS2}
\end{equation}
Combining Eqs.~\eqref{eq:thm4-triangle} and~\eqref{eq:thm4-LS2},
\[
  \|\ehat - e\|
  \;\le\; \sqrt{\Cadapt}\,\non^{-\beta/(2\beta+k)}
  + \siglab/\sqrt{\non} + \estar.
\]
Applying Proposition~\ref{prop:mc-refinement} yields the claim. \qed

\subsection{Proof of Theorem~\ref{thm:combined}\textup{(i)}}\label{app:proof-thm5}

The bound of Theorem~\ref{thm:combined}\textup{(i)} is that of Theorem~\ref{thm:est-error} with the term $\termRB$ replaced by $\termRMC + \termW$.
Since $\termRB$ is $\kappa_B/(1-\gamma)$ times the binned refinement error $\|(I - \PiEB)e\|$, it suffices to bound this error.
The residual $(I - \PiE)e$ is orthogonal in $L^2(\rho)$ to every function of $\ehat$, and $(\PiE - \PiEB)e$ is a function of $\ehat$, so
\begin{equation}
  \|(I - \PiEB)\,e\|^2
  \;=\; \|(I - \PiE)\,e\|^2
  + \|(\PiE - \PiEB)\,e\|^2.
  \label{eq:thm5-pyth}
\end{equation}
The first term is the squared refinement error of Eq.~\eqref{eq:refinement-error}. For the second, let $m_e(x) := \E_\rho[e \mid \ehat(S) = x]$, taken as
the version of total variation at most $\CTV$, so that $\PiE e = m_e \circ \ehat$
and, by the tower property, $\PiEB e$ equals the $\rho$-average of
$m_e \circ \ehat$ on each bin. Let $\omega_j$ be the oscillation of $m_e$ on
the bin $I_j$. A random variable confined to an interval of length $\omega_j$
has variance at most $\omega_j^2/4$, and $\sum_j \omega_j \le \CTV$ because the
bins are disjoint intervals, so
\begin{align*}
  \|(\PiE - \PiEB)\,e\|^2
  &= \sum_{j=1}^B \rho(\ehat(S) \in I_j)\,
    \mathrm{Var}_\rho\big(m_e(\ehat(S)) \,\big|\, \ehat(S) \in I_j\big)\\
  &\le \frac{c_{\mathrm{bin}}}{4B}\Big(\sum_{j=1}^B \omega_j\Big)^2
  \le \frac{c_{\mathrm{bin}} \CTV^2}{4B}.
\end{align*}
Hence $\|(\PiE - \PiEB)e\| \le \tfrac12 \CTV \sqrt{c_{\mathrm{bin}}/B}$.
By Theorem~\ref{thm:rate-transfer}, with probability at least $1 - \delta$, the refinement error $\|(I - \PiE)e\|$ is at most the bracketed sum in the definition of $\termRMC$.
On this event, Eq.~\eqref{eq:thm5-pyth} and the inequality $\sqrt{a^2 + b^2} \le a + b$ for $a, b \ge 0$ give
\[
  \termRB \;\le\; \frac{\kappa_B}{1-\gamma}\Big(\|(I - \PiE)\,e\| + \|(\PiE - \PiEB)\,e\|\Big)
  \;\le\; \termRMC + \termW.
\]
The bound of Theorem~\ref{thm:est-error} holds with probability at least $1 - \delta$, so both hold with probability at least $1 - 2\delta$. \qed

\section{Auxiliary Results}\label{app:aux}

\subsection{Histogram calibration error}\label{app:thm2}

Throughout this appendix, $P_0$ denotes the distribution of an off-policy transition $O = (S, A, R, S')$, $P_n$ the empirical distribution of the calibration sample $\Doff$, and $Pf := \int f \, dP$ for a distribution $P$; when $f$ depends on $\Doff$, $P_0 f$ integrates over an independent observation $O$ with $f$ held fixed.
For a function $f$ of $O$, we write $\|f\|^2 := P_0 f^2$.
For a function $f$ of $(S, A)$, this is the norm $\|f\|_{\rho,b}$ of Section~\ref{sec:finite-sample}.
The DR target $\Tpihat$ in Eq.~\eqref{eq:dr-bellman} has the state-level version $\Tzeropihat(v)(s) := \E[\Tpihat(v)(S, A, R, S') \mid S = s]$, where the expectation is over the true conditional law of $(A, R, S')$ given $S = s$ with the nuisance estimates held fixed.
By the DR error identity of \citet[Theorem~2]{vanderlaan2025bc}, $\Tzeropihat(v)(s) - \Tpi(v)(s) = \sum_a b(a \mid s)\,(\wpi - \wpihat)(a \mid s)\,(\hat q_v - q_v)(s, a)$, where $q_v(s, a) := r(s, a) + \gamma (P v)(s, a)$.
All finite-sample results below take the partition $I_1,\ldots,I_B$ as fixed and assume $1\le B\le\noff/2$.
\paragraph{Conditions.}
\begin{description}
  \item[C1. Boundedness.] $R$, $r$, $\hat{r}$, $\wpi$, $\wpihat$, $\vhat$,
    $\ehat$ are uniformly bounded by some $M \in (0, \infty)$.
    For the analyzed horizon $K$, also assume $\max_{0\le k\le K}\|\hat v^{(k)}\|_\infty\le M$ almost surely.
    The same deterministic $M$ is independent of $K$, $B$, and the sample sizes.
  \item[C2. Independent sampling and sample splitting.]
    $\Doff=\{O_i\}_{i=1}^{\noff}$, where $O_i = (S_i, A_i, R_i, S_i')$, consists of i.i.d. observations from $P_0$ and is independent of all training data used to obtain $\hat r,\wpihat,\hat P,\vhat,\ehat$.
  \item[C3. Bounded linear transition estimate.] $\hat P$ is linear and
    $\|\hat P f\|_\infty\le L\|f\|_\infty$ for every bounded ${f : \mathcal{S} \to \mathbb{R}}$, with a fixed $L<\infty$.
    A fitted Markov transition kernel satisfies this with $L=1$.
\end{description}
The iterate bound in C1 is an additional stability assumption on the analyzed runs, not a consequence of bounded inputs or importance weights.
Conditions C1 and C3 imply a common bound on the DR targets for all $v\in\vhat+\FBE$ with $\|v\|_\infty\le M$.
Constants in the finite-sample results may depend on $M$ and $L$, but not on $K$, $B$, or the sample sizes; we write $x\lesssim y$ when $x\le c\,y$ for such a constant $c$.
By C2, conditional on all training data, these estimators are fixed and the calibration observations remain i.i.d. from $P_0$.
The probability bounds below are first obtained conditionally and then averaged over training data.

For a predictor $v$, the binned Bellman calibration error is $\CalGBCB(v) := \|\PiEB[\Tpi(v) - v]\|$, with $\PiEB$ as in Section~\ref{sec:finite-sample}.

\begin{theorem}[Histogram calibration error]\label{thm:hist-cal}
Under C1, C2, C3, there is a constant $C \in (0, \infty)$ such that,
for any $K \in \mathbb{N}$ and with probability at least $1 - \delta$,
\begin{align*}
  \CalGBCB(\vK)
  &\le C\!\left[\sqrt{\tfrac{B}{\noff}\log\tfrac{\noff}{B}}
    + \sqrt{\tfrac{\log(1/\delta)}{\noff}}\right]\\
  &\quad + \big\|(\wpihat - \wpi)(\hat q_{\vK} - q_{\vK})\big\|
  + \big\|\Tzeropihat(\vK) - \Tzeropihat(\hat v^{(K-1)})\big\|.
\end{align*}
\end{theorem}

The last term is the change between the last two iterates under
$\Tzeropihat$; it is left explicit.

\begin{proof}
The argument follows the proof of Theorem~3 of \citet{vanderlaan2025bc} with the
conditioning axis replaced by $\ehat$ and the additive structure of
$\hat v^{(k)} = \vhat + \theta^{(k)}\circ\ehat$ accounted for in the
empirical-process step.

\emph{Step 1 (First-order optimality).} The histogram regression of pseudo-residuals
$\hat\xi_i^{(K-1)} = \Tpihat(\hat v^{(K-1)})(O_i) - \vhat(S_i)$ on the bins of
$\ehat(S_i)$ satisfies binwise normal equations. Substituting
$\vK(S_i) = \vhat(S_i) + \theta_n^{(K)}(\ehat(S_i))$ and using linearity, for any
$f \in \FB$,
\begin{equation}
  \tfrac{1}{\noff}\sum_i f(\ehat(S_i))\big[\vK(S_i)
    - \Tpihat(\hat v^{(K-1)})(O_i)\big] \;=\; 0.
  \label{eq:thm2-foc}
\end{equation}

\emph{Step 2 (Inner-product reduction).} Set $g := \PiEB[\vK - \Tpi(\vK)]$, so
$\|g\| = \CalGBCB(\vK)$ and $g \in \FBE$. Self-adjointness of $\PiEB$ in
$L^2(\rho)$ and $\PiEB g = g$ yield
\[
  \|g\|^2 \;=\; P_0\!\big[g\cdot(\vK - \Tpi(\vK))\big].
\]

\emph{Step 3 (Decomposition).} Adding and subtracting $\Tzeropihat(\hat v^{(K-1)})$
and $\Tzeropihat(\vK)$,
\begin{align*}
  \|g\|^2 &= P_0\big[g\cdot(\vK - \Tzeropihat(\hat v^{(K-1)}))\big]
    \tag{(a)}\\
  &\quad + P_0\big[g\cdot(\Tzeropihat(\hat v^{(K-1)})
    - \Tzeropihat(\vK))\big] \tag{(b)}\\
  &\quad + P_0\big[g\cdot(\Tzeropihat(\vK) - \Tpi(\vK))\big]. \tag{(c)}
\end{align*}
By the DR error identity (Theorem~2 of \citet{vanderlaan2025bc}) and
Cauchy-Schwarz,
$|(c)| \le \|g\|\cdot\|(\wpihat - \wpi)(\hat q_{\vK} - q_{\vK})\|$, and
$|(b)| \le \|g\|\cdot\|\Tzeropihat(\vK) - \Tzeropihat(\hat v^{(K-1)})\|$.

\emph{Step 4 (Empirical process for (a)).} Since $\Tpihat(v)(O)$ is an
unbiased estimator of $\Tzeropihat(v)$ given $S$, and Eq.~\eqref{eq:thm2-foc} with
$f$ chosen so $f\circ\ehat = g$ gives $P_n[g\cdot(\vK - \Tpihat(\hat v^{(K-1)}))]
= 0$,
\[
  |(a)| \;=\; \big|(P_n - P_0)[g\cdot(\vK - \Tpihat(\hat v^{(K-1)}))]\big|.
\]
Let $\mathcal V_M:=\{v\in\vhat+\FBE:\|v\|_\infty\le M\}$.
The integrand belongs to the bounded function class
\[
  \mathcal{H}_{\GBC}:=\{h\,[v-\Tpihat(u)]:
      h\in\FBE,\ \|h\|_\infty\le5M,\ u,v\in\mathcal V_M\}.
\]
By C1, $g$ and both iterates lie in the indicated bounded classes.
Conditional on training, the first factor is a bounded $B$-dimensional histogram class.
The second factor is a bounded affine class of dimension at most $2B$, because the DR target is affine in its argument under C3.
Uniform covering bounds for bounded finite-dimensional classes, followed by a product cover, therefore give
$\sup_Q\log N(\varepsilon,\mathcal H_{\GBC},L_2(Q))\lesssim B\log(c_0/\varepsilon)$,
where $c_0$ depends only on the common bounds.
The entropy integral up to radius $\tau$ is consequently of order $\tau\sqrt{B\log(c_0/\tau)}$.

\emph{Step 5 (Local maximal inequality).} Let
$\delta_n := \sqrt{(B/\noff)\log(\noff/B)}$; by Step~4, the critical radius of $\mathcal{H}_{\GBC}$ is of this order.
If $\delta_n$ exceeds the envelope scale, the stated bound follows directly from boundedness; otherwise the local maximal inequality of \citet[Lemma~3]{vanderlaan2025bc} applies, since $\mathcal{H}_{\GBC}$ is uniformly bounded and star-shaped (scaling $h$ by a factor in $[0,1]$ stays in the class).
With probability at least $1 - \delta$,
every $\zeta \in \mathcal{H}_{\GBC}$ satisfies
\[
  |(P_n - P_0)\zeta|
  \;\lesssim\; \delta_n^2 + \delta_n\|\zeta\|
  + \tfrac{\sqrt{\log(1/\delta)}}{\sqrt{\noff}}\,\|\zeta\|
  + \tfrac{\log(1/\delta)}{\noff}.
\]
The event is uniform over $\mathcal{H}_{\GBC}$, so it covers the data-dependent $g$ and iterates.
Specializing to $\zeta = g\cdot(\vK - \Tpihat(\hat v^{(K-1)}))$ with $\|\zeta\|
\lesssim \|g\|$ and combining (a)-(c) gives $\|g\|^2 \le a_n + \|g\|\,b_n$,
where, for a constant $c$,
\begin{align*}
  a_n &:= c\,\big[\delta_n^2 + \log(1/\delta)/\noff\big],\\
  b_n &:= c\,\big[\delta_n + \sqrt{\log(1/\delta)/\noff}\big]
    + \big\|(\wpihat - \wpi)(\hat q_{\vK} - q_{\vK})\big\|\\
  &\qquad + \big\|\Tzeropihat(\vK) - \Tzeropihat(\hat v^{(K-1)})\big\|.
\end{align*}
Solving the quadratic inequality gives $\|g\| \le b_n + \sqrt{a_n}$, and
$\sqrt{a_n} \le \sqrt{c}\,\big[\delta_n + \sqrt{\log(1/\delta)/\noff}\big]$
yields the claim.
\end{proof}

\subsection{Inexact iteration and per-iteration error}\label{app:lemmas-78}

\begin{description}
  \item[C4. Stationary coverage.] There exists a stationary probability measure
    $\mu_B$ for $\PpiEB$ and $\kappa_B \in (0, \infty)$ such that
    $\|h\|_{\mu_B} \le \kappa_B \|h\|$ for all
    $h : \mathcal{S} \to \mathbb{R}$. Under $\rho \Ppi = \rho$, $\mu_B
    = \rho$ and $\kappa_B = 1$ are admissible.
\end{description}

The binned operator $\TGBCB$ of Section~\ref{sec:finite-sample}
is a $\gamma$-contraction in $\|\cdot\|_\infty$ (the proof transfers verbatim
from Lemma~\ref{lem:fixed-point}), so it has a unique bounded fixed point
$\vzeroGBCB$, and $\vzeroGBCB - \vhat \in \FBE$. Under C4, $\PpiEB$ is
non-expansive in $\|\cdot\|_{\mu_B}$, so $\TGBCB$ is a
$\gamma$-contraction in $\|\cdot\|_{\mu_B}$.

\begin{lemma}[Inexact iterations of $\TGBCB$]\label{lem:inexact-iter}
Assume C4. Let $\{\eta_k\}_{k\ge 1}$ be any sequence with
$\|\hat v^{(k)} - \TGBCB(\hat v^{(k-1)})\|_{\mu_B} \le \eta_k$ for all
$k \ge 1$. Then for any $K \ge 1$,
\[
  \|\vK - \vzeroGBCB\|_{\mu_B}
  \;\le\; \gamma^K\,\|\hat{v}^{(0)} - \vzeroGBCB\|_{\mu_B}
  + \sum_{j=1}^K \gamma^{K-j}\,\eta_j.
\]
\end{lemma}

\begin{proof}
Set $\Delta_k := \hat v^{(k)} - \vzeroGBCB$
and $\psi^{(k)} := \hat v^{(k)} - \TGBCB(\hat v^{(k-1)})$, so
$\|\psi^{(k)}\|_{\mu_B} \le \eta_k$. Using
$\TGBCB(\vzeroGBCB) = \vzeroGBCB$ and the linear contraction
$\TGBCB(g_1) - \TGBCB(g_2) = \gamma\PpiEB(g_1 - g_2)$,
\[
  \Delta_k \;=\; \psi^{(k)} + \gamma\,\PpiEB \Delta_{k-1}.
\]
Triangle inequality and non-expansion of $\PpiEB$ in
$\|\cdot\|_{\mu_B}$ give
$\|\Delta_k\|_{\mu_B} \le \eta_k + \gamma\|\Delta_{k-1}\|_{\mu_B}$.
Unrolling from $k = K$ down to $k = 1$ with $\Delta_0
= \hat{v}^{(0)} - \vzeroGBCB$ yields the claim.
\end{proof}

\begin{lemma}[Per-iteration error]\label{lem:per-iter-error}
Under C1, C2, C3, there is a constant $C < \infty$ such that, with
probability at least $1 - \delta$, uniformly over $1 \le k \le K$,
\[
  \|\hat v^{(k)} - \TGBCB(\hat v^{(k-1)})\|
  \;\le\; C\!\left[\sqrt{\tfrac{B}{\noff}\log\tfrac{\noff}{B}}
  + \sqrt{\tfrac{\log(K/\delta)}{\noff}}\right]
  + \big\|(\wpihat - \wpi)(\hat q_{\hat v^{(k-1)}} - q_{\hat v^{(k-1)}})\big\|.
\]
\end{lemma}

\begin{proof}
Let $\psi^{(k)} := \hat v^{(k)} - \TGBCB(\hat v^{(k-1)})$. Then
$\psi^{(k)} = \theta_n^{(k)}\circ\ehat - \PiEB[\Tpi(\hat v^{(k-1)}) - \vhat]
\in \FBE$, since both summands lie in $\FBE$ and the offset $\vhat$ cancels.

\emph{Step 1 (Inner-product reduction).} Self-adjointness of $\PiEB$ and
$\PiEB \psi^{(k)} = \psi^{(k)}$ give, for any $g \in L^2(\rho)$,
$P_0[\psi^{(k)}\cdot \PiEB g] = P_0[\psi^{(k)}\cdot g]$. Applied to
$g = \Tpi(\hat v^{(k-1)}) - \vhat$ inside the expansion of $\TGBCB$,
\[
  \|\psi^{(k)}\|^2
  \;=\; P_0\big[\psi^{(k)}\cdot(\hat v^{(k)} - \Tpi(\hat v^{(k-1)}))\big].
\]

\emph{Step 2 (DR decomposition).} Adding and subtracting
$\Tzeropihat(\hat v^{(k-1)})$,
\begin{align*}
  \|\psi^{(k)}\|^2
  &= P_0[\psi^{(k)}\cdot(\hat v^{(k)} - \Tzeropihat(\hat v^{(k-1)}))]
    \tag{(a)}\\
  &\quad + P_0[\psi^{(k)}\cdot(\Tzeropihat(\hat v^{(k-1)})
    - \Tpi(\hat v^{(k-1)}))]. \tag{(b)}
\end{align*}
By the DR error identity and Cauchy-Schwarz, $|(b)| \le \|\psi^{(k)}\|\cdot
\|(\wpihat - \wpi)(\hat q_{\hat v^{(k-1)}} - q_{\hat v^{(k-1)}})\|$.

\emph{Step 3 (Empirical process for (a)).} The first-order optimality condition
\eqref{eq:thm2-foc} (with $K$ replaced by $k$ and $f\circ\ehat = \psi^{(k)}$) gives
$P_n[\psi^{(k)}\cdot(\hat v^{(k)} - \Tpihat(\hat v^{(k-1)}))] = 0$, and
$P_0[\psi^{(k)}\cdot \Tpihat(\hat v^{(k-1)})] =
P_0[\psi^{(k)}\cdot\Tzeropihat(\hat v^{(k-1)})]$ by the law of total expectation,
hence $|(a)| = |(P_n - P_0)[\psi^{(k)}\cdot(\hat v^{(k)}
- \Tpihat(\hat v^{(k-1)}))]|$.

\emph{Step 4 (Entropy and union bound).} The integrand lies in
$\mathcal{H}_{\GBC}$ (Theorem~\ref{thm:hist-cal} proof, Step~4), since
$\|\psi^{(k)}\|_\infty\le(4+\gamma)M\le5M$.
The entropy bound and local maximal inequality apply with confidence parameter
$\delta/K$ to allow a union bound over $1 \le k \le K$. With probability at least
$1 - \delta$ uniformly over $k$,
$|(a)| \lesssim \delta_n^2 + \delta_n\|\psi^{(k)}\| +
\sqrt{\log(K/\delta)/\noff}\,\|\psi^{(k)}\| + \log(K/\delta)/\noff$, with
$\delta_n$ as in the proof of Theorem~\ref{thm:hist-cal}.

Combining (a) and (b) on this event gives, for every $1 \le k \le K$,
$\|\psi^{(k)}\|^2 \le a_n + \|\psi^{(k)}\|\,b_{n,k}$, where, for a constant $c$,
\begin{align*}
  a_n &:= c\,\big[\delta_n^2 + \log(K/\delta)/\noff\big],\\
  b_{n,k} &:= c\,\big[\delta_n + \sqrt{\log(K/\delta)/\noff}\big]
    + \big\|(\wpihat - \wpi)(\hat q_{\hat v^{(k-1)}} - q_{\hat v^{(k-1)}})\big\|.
\end{align*}
Solving the quadratic inequality gives
$\|\psi^{(k)}\| \le b_{n,k} + \sqrt{a_n}$, and
$\sqrt{a_n} \le \sqrt{c}\,\big[\delta_n + \sqrt{\log(K/\delta)/\noff}\big]$
yields the claim.
\end{proof}

\paragraph{Proof of Theorem~\ref{thm:est-error}.}
Let $C$ be the constant of Lemma~\ref{lem:per-iter-error} and, for $1 \le k \le K$,
set
\[
  \eta_k := \kappa_B\Big(
    C\Big[\sqrt{\tfrac{B}{\noff}\log\tfrac{\noff}{B}}
    + \sqrt{\tfrac{\log(K/\delta)}{\noff}}\Big]
    + \big\|(\wpihat - \wpi)(\hat q_{\hat v^{(k-1)}} - q_{\hat v^{(k-1)}})\big\|
  \Big).
\]
On the event of Lemma~\ref{lem:per-iter-error}, which has probability at least
$1 - \delta$, C4 gives, for every $1 \le k \le K$,
\[
  \|\hat v^{(k)} - \TGBCB(\hat v^{(k-1)})\|_{\mu_B}
  \;\le\; \kappa_B\,\|\hat v^{(k)} - \TGBCB(\hat v^{(k-1)})\|
  \;\le\; \eta_k.
\]
Lemma~\ref{lem:inexact-iter} and
$\sum_{j=1}^K \gamma^{K-j} \le 1/(1-\gamma)$ then give
\begin{align*}
  \|\vK - \vzeroGBCB\|_{\mu_B}
  &\le \gamma^K\,\|\hat{v}^{(0)} - \vzeroGBCB\|_{\mu_B}
    + \sum_{j=1}^K \gamma^{K-j}\,\eta_j\\
  &\le \gamma^K\,\|\hat{v}^{(0)} - \vzeroGBCB\|_{\mu_B}
    + \frac{1}{1-\gamma}\,\max_{1 \le j \le K}\eta_j.
\end{align*}
The argument of Appendix~\ref{app:proof-thm1}, with $\PiE$ and $\mu$
replaced by $\PiEB$ and $\mu_B$, together with C4 gives
\[
  \|\vzeroGBCB - \vzero\|_{\mu_B}
  \;\le\; \frac{1}{1-\gamma}\,\|(I - \PiEB)\,e\|_{\mu_B}
  \;\le\; \frac{\kappa_B}{1-\gamma}\,\|(I - \PiEB)\,e\|.
\]
The triangle inequality
$\|\vK - \vzero\|_{\mu_B} \le \|\vK - \vzeroGBCB\|_{\mu_B}
+ \|\vzeroGBCB - \vzero\|_{\mu_B}$ combines the last two displays
into the bound of Theorem~\ref{thm:est-error}. \qed

\subsection{Proof of Theorem~\ref{cor:rate-transfer-td1}}\label{app:cor41}

The triangle inequality
$\|\ehat - (I - \gamma\Ppi)e\| \le \|\ehat - g\circ\phi\| + \|g\circ\phi
- (I - \gamma\Ppi)e\|$ together with Assumption~\ref{ass:LS1} for
$f^* = (I - \gamma\Ppi)e$ and Assumption~\ref{ass:LS2} gives, on the
high-probability event,
\[
  \|\ehat - (I - \gamma\Ppi)e\|
  \;\le\; \sqrt{\Cadapt}\,\non^{-\beta/(2\beta+k)}
  + \siglab/\sqrt{\non} + \estar.
\]
Substituting this bound into
Proposition~\ref{prop:td1-refinement} yields the claim. \qed

\subsection{Proof of Theorem~\ref{thm:combined}\textup{(ii)}}\label{app:cor51}

The proof of case (i) (Appendix~\ref{app:proof-thm5}) applies with one change.
The refinement error $\|(I - \PiE)e\|$ is bounded by Theorem~\ref{cor:rate-transfer-td1} in place of Theorem~\ref{thm:rate-transfer}, again with probability at least $1 - \delta$.
Multiplying the bound of Theorem~\ref{cor:rate-transfer-td1} by $\kappa_B/(1-\gamma)$ gives
\[
  \kappa_B\sqrt{\Cocc}\left[
  \frac{\sqrt{\Cadapt}\,\non^{-\beta/(2\beta+k)}
  + \siglab/\sqrt{\non} + \estar}{(1-\gamma)^2}
  + \frac{\gamma\,\BEhat\,\|\ehat\|_\infty}{(1-\gamma)^3}\right],
\]
which is the term $\termRTD$. \qed

\section{Implementation Notes and Additional Experiments}\label{app:c}

\subsection{Implementation notes}\label{app:impl}

This subsection collects practical guidance that the main text defers.

\paragraph{Number of bins $B$.}
We use $B = 32$ in every environment. The
bound of Theorem~\ref{thm:combined} balances its terms $\termW$ and $\termS$ at $B \asymp (\noff/\log\noff)^{1/2}$ (Section~\ref{sec:finite-sample});
with the proportionality constant set to one, this reference value is roughly
$25$ to $95$ for the calibration samples in our experiments
($\noff$ from $5\times 10^3$ to about $10^5$ transitions), although the
constant itself is not determined by the theory. We report the sensitivity
to the bin count in Section~\ref{app:additional-exp-details}. The bin edges are
the empirical quantiles of $\ehat$ on $\Doff$, so the
bins hold roughly equal numbers of points.

\paragraph{Base predictor $\vhat$.}
The Toy MDP base predictor is a linear model, fit to the value of the behavior policy by fitted value iteration ($40$ iterations, L2 regularization coefficient $2 \times 10^{-5}$), and intentionally misspecified because the reward depends nonlinearly on $x_2$. CRM uses three classes of base predictor in parallel: a linear model with the same L2 regularization as on Toy (linear), a one-hidden-layer torch MLP with $32$ units, dropout $0.1$, Adam at learning rate $10^{-4}$ trained for $10$ epochs per fitting iteration with batch size $1024$ (neural), and a gradient-boosted regressor with $50$ trees of depth $2$ at learning rate $0.03$ (boosted). For the main CRM results, all three classes are fit to the value of the behavior policy by fitted value iteration. The change under importance-weighted fitting is reported in Section~\ref{app:additional-exp-details}. The D4RL base predictor is a two-hidden-layer MLP with $64$ units per layer, trained with the settings of the error model below by Monte-Carlo regression of returns (no fitted value iteration) on $2{,}000$ state-return pairs from $50$-step windows of \texttt{simple-v0} trajectories.

\paragraph{Error model $\ehat$.}
On Toy, CRM, and D4RL, the error model is an MLP with two hidden layers of $64$ ReLU units, trained by Adam at learning rate $10^{-3}$ with minibatches of up to $200$ for at most $200$ epochs, with L2 regularization coefficient $5 \times 10^{-6}$. By default, $\ehat$ is obtained by TD training, fit by least squares on on-policy transitions from $\Don$ with training target $r + \gamma\,\vhat(s') - \vhat(s)$. This corresponds to the estimand $f^\star = (I - \gamma\Ppi)e$ analyzed in Proposition~\ref{prop:td1-refinement}. MC training with full returns has estimand $e$, as analyzed in Proposition~\ref{prop:mc-refinement}. The Direct baseline outputs $\vhat + \ehat$.

\paragraph{Iteration count $K$.}
We use $K = 3$ as the default in every environment. Theorem~\ref{thm:combined} admits $K \asymp \log\noff$ up to a constant that the theory does not determine.

\paragraph{Data and policies.}
On the Toy MDP the state is $s \in [0, 1]^{d}$ with $d = 10$, there are two actions, and $\gamma = 0.8$; the target and behavior policies take action $1$ with probability $1/(1 + e^{-\beta(x_1 - 0.5)})$, with $\beta = 3$ and $\beta = 1$ respectively. The reward is $x_1(1 + 0.5a) + 1.5\sin(2\pi x_2)\,a + \varepsilon$ with $\varepsilon \sim \mathcal{N}(0, 0.1^2)$. The next state has $x_1' = \mathrm{clip}(0.5x_1 + 0.25a + 0.25 + \xi_1)$ and $x_2' = \mathrm{clip}(0.5x_2 + 0.1a + 0.25 + \xi_2)$, where $\xi_1, \xi_2 \sim \mathcal{N}(0, 0.1^2)$ and clip truncates to $[0, 1]$, while $x_3', \dots, x_d'$ are drawn independently from $U[0.4, 0.6]$. Initial states have $x_1, x_2 \sim U[0, 1]$ and $x_3, \dots, x_d \sim U[0.4, 0.6]$. The off-policy data are $10{,}000$ transitions from $1{,}000$ behavior episodes of length $10$; $500$ episodes train $\vhat$ and the other $500$ form $\Doff$ ($\noff = 5{,}000$), and $\Don$ is $100$ transitions from $10$ target-policy episodes. On CRM, the target policy is a deterministic promotion rule smoothed with probability $10^{-3}$, and the behavior policy is an equal mixture of the target policy and a heuristic rule. The two rules are based on the target and behavior policies described by \citet{vanderlaan2025bc}. The off-policy data are $15{,}000$ behavior customers observed for up to $24$ months: $5{,}000$ train $\vhat$ (about $5 \times 10^4$ transitions) and the other $10{,}000$ form $\Doff$ (about $10^5$ transitions), and $\Don$ is $1{,}000$ target-policy customers (about $10^4$ transitions). On D4RL, the behavior and target policies are the released checkpoints of the policies that generated \texttt{medium-v0} and \texttt{expert-v0} and $\gamma = 0.9$; $\Doff$ is ten $10$-step windows from each of $100$ \texttt{medium-v0} episodes (about $10^4$ transitions), and $\Don$ is one $100$-step window from each of $10$ \texttt{expert-v0} episodes (about $10^3$ transitions).

\paragraph{Calibration sample and DR target.}
The DR target uses the known ratio $\wpi = \pi/b$, computed from the simulator's policy probabilities on Toy and CRM and from the policy checkpoints on D4RL. We obtain $\wpihat$ by clipping $\wpi$ to $[0.1, 10]$ and dividing it by its mean over $\Doff$; the on-policy transitions appended to BC's calibration data keep weight one. For a calibration transition $(S_i, A_i, R_i, S_i')$ and the current iterate $\hat{v}^{(k)}$, the target is $\hat{v}^{(k)}(S_i) + \wpihat(A_i \mid S_i)\{R_i + \gamma\,\hat{v}^{(k)}(S_i') - \hat{v}^{(k)}(S_i)\}$, which is Eq.~\eqref{eq:dr-bellman} with $\hat{q}_v(s, a) := v(s)$, so we do not fit a reward or transition model. Every iteration uses all of $\Doff$, as in Algorithm~\ref{alg:gbc}.

\paragraph{Ground-truth value function.}
The ground truth $\vzero$ is computed by Monte Carlo. On the Toy MDP, we evaluate on a $10\times 10$ grid in $(x_1, x_2)$, with $x_3, \dots, x_d$ set to $0.5$, using $1500$ rollouts of horizon $31$ each. On CRM we use $300$ initial states with $30$ rollouts to absorption (eval horizon $60$ months), and we report RMSE multiplied by $1 - \gamma$, as in \citet{vanderlaan2025bc}. On D4RL we sample $500$ states from the held-out $10\%$ of target trajectories and use the per-state Monte-Carlo return-to-go on its parent trajectory. Throughout, SE is the sample standard deviation over seeds divided by the square root of the number of seeds.

\subsection{Additional experiments and diagnostics}\label{app:additional-exp}

\subsubsection{Conditioning-axis diagnostic}\label{app:origin-diagnostic}

On the runs of Table~\ref{tab:crm-headline}, we compute the true prediction error $e = \vzero - \vhat$ on the evaluation states of Appendix~\ref{app:impl}.

\paragraph{Axis correlations.}
Table~\ref{tab:origin-diag} reports the empirical Pearson correlation of each candidate conditioning axis with the true prediction error and the within-bin standard-deviation reduction achieved by histogram averaging on that axis. Binning on $\ehat$ with $B = 32$ reduces the within-bin standard deviation of $e$ more than binning on $\vhat$ for the boosted and neural predictors ($64\%$ against $49\%$ and $36\%$), and by a similar amount for the linear predictor ($52\%$ against $50\%$ with $8$ fitting iterations, and $30\%$ for both with $18$). The correlation with $e$ is higher for $\ehat$ in all four settings.

\begin{table}[h]
\centering
\caption{Conditioning-axis diagnostic on CRM (mean $\pm$ SE over $10$ seeds). Within-bin std reduction is one minus the ratio of the mean of the per-bin standard deviations of $e$ to its overall standard deviation, at $B = 32$.}
\label{tab:origin-diag}
\small
\begin{tabular}{lcccc}
\toprule
 & \multicolumn{2}{c}{$\mathrm{corr}(\text{axis}, e)$} & \multicolumn{2}{c}{Within-bin std reduction} \\
\cmidrule(lr){2-3}\cmidrule(lr){4-5}
Setting & $\vhat$ (BC) & $\ehat$ (\GBC{}) & $\vhat$ (BC) & $\ehat$ (\GBC{}) \\
\midrule
Linear, $8$ iter.   & $+0.75 \pm 0.01$ & $+0.81 \pm 0.01$ & $50 \pm 1\%$ & $52 \pm 1\%$ \\
Linear, $18$ iter.  & $+0.34 \pm 0.02$ & $+0.58 \pm 0.02$ & $30 \pm 1\%$ & $30 \pm 1\%$ \\
Boosted, $18$ iter. & $+0.73 \pm 0.02$ & $+0.88 \pm 0.01$ & $49 \pm 2\%$ & $64 \pm 1\%$ \\
Neural, $100$ iter. & $+0.64 \pm 0.05$ & $+0.88 \pm 0.01$ & $36 \pm 3\%$ & $64 \pm 1\%$ \\
\bottomrule
\end{tabular}
\end{table}

\paragraph{Comparing axes without ground truth.}
The true prediction error $e$ is not available in practice. The two axes can instead be compared on data not used to fit them. First, cross-fit $\ehat$ on $\Don$ and inspect its held-out error on the training targets and its learning curve over subsamples of $\Don$. Then, on untouched folds of the off-policy sample, compare how well $\ehat$-bins and $\vhat$-bins predict a doubly robust estimate of the current predictor's one-step Bellman residual. A stable reduction in held-out prediction error with $\ehat$-bins supports using \GBC{}. With TD training, the next-error-bin diagnostic (Appendix~\ref{app:next-bin-diagnostic}) additionally compares how well the next error bin is predicted from the current bin with and without the full state.

\subsubsection{Next-error-bin prediction diagnostic}\label{app:next-bin-diagnostic}

With TD training, the bound of Proposition~\ref{prop:td1-refinement} contains the inherent Bellman error $\BEhat$ (Definition~\ref{def:ibe-ehat}), which is zero when the conditional distribution of $\ehat(S')$ given $S$ depends on $S$ only through $\ehat(S)$. As a proxy for $\BEhat$, this diagnostic measures how much the prediction of the next error bin degrades when only the current error bin is used instead of the full state. We divide the values of $\ehat$ into $B = 32$ quantile bins and predict the bin of $\ehat(S')$ in two ways, from the bin of $\ehat(S)$ alone and from that bin together with the full state $S$ using an ExtraTrees classifier. In each run of Tables~\ref{tab:crm-headline} and~\ref{tab:d4rl-headline}, we fit both predictors on about $10^4$ target-policy transitions not used in those runs and compare them by the multiclass Brier score on another $10^4$ such transitions.

Table~\ref{tab:next-bin} reports the two scores. On CRM, the bins alone achieve $80\%$ to $99\%$ of the reduction in Brier score obtained with the full state, measured from a uniform prediction. On D4RL, they achieve $31\%$ to $40\%$. The larger degradation on D4RL may be one factor behind the smaller improvement of \GBC{} over Raw in Table~\ref{tab:d4rl-headline}.

\begin{table}[h]
\centering
\caption{Next-error-bin prediction diagnostic: held-out multiclass Brier score for predicting the next error bin (mean $\pm$ SE over $10$ seeds; lower is better). A uniform prediction scores about $0.97$.}
\label{tab:next-bin}
\small
\begin{tabular}{llcc}
\toprule
Environment & Setting & Bin only & Bin and state \\
\midrule
CRM  & Linear, $8$ iter.    & $0.779 \pm 0.004$ & $0.744 \pm 0.005$ \\
     & Linear, $18$ iter.   & $0.826 \pm 0.004$ & $0.791 \pm 0.005$ \\
     & Boosted, $18$ iter.  & $0.793 \pm 0.003$ & $0.792 \pm 0.003$ \\
     & Neural, $100$ iter.  & $0.787 \pm 0.004$ & $0.785 \pm 0.004$ \\
\midrule
D4RL & HalfCheetah        & $0.886 \pm 0.011$ & $0.760 \pm 0.015$ \\
     & Hopper             & $0.894 \pm 0.004$ & $0.762 \pm 0.006$ \\
     & Walker2d           & $0.942 \pm 0.002$ & $0.880 \pm 0.004$ \\
\bottomrule
\end{tabular}
\end{table}

\FloatBarrier
\subsubsection{Additional experimental results}\label{app:additional-exp-details}

As in Section~\ref{sec:experiments}, bold in the RMSE tables below marks, in each column, the method with the lowest mean RMSE and any method whose paired difference from it is within two SE.

\paragraph{Toy MDP: sweeps over $K$, $B$, and $\non$.}
Tables~\ref{tab:toy-iter}, \ref{tab:toy-bins}, and~\ref{tab:toy-non} vary $K$, $B$, and $\non$ one at a time from the defaults ($K = 3$, $B = 32$, $\non = 100$) on the seeds of Table~\ref{tab:toy-headline}. Within each seed, the off-policy data, $\vhat$, and the evaluation grid are the same in every setting, so Raw is $0.429$ throughout.

\begin{table}[h]
\centering
\caption{Toy MDP RMSE by iteration count $K$ ($\non = 100$, $B = 32$; mean $\pm$ SE over $10$ seeds).}
\label{tab:toy-iter}
\small
\begin{tabular}{lccccc}
\toprule
Method & $K{=}1$ & $K{=}3$ & $K{=}5$ & $K{=}10$ & $K{=}20$ \\
\midrule
BC     & $0.484 \pm 0.007$ & $0.451 \pm 0.006$ & $0.419 \pm 0.005$ & $0.392 \pm 0.005$ & $0.386 \pm 0.004$ \\
\GBC{} & $\mathbf{0.388 \pm 0.008}$ & $\mathbf{0.364 \pm 0.010}$ & $\mathbf{0.352 \pm 0.012}$ & $\mathbf{0.347 \pm 0.014}$ & $\mathbf{0.349 \pm 0.014}$ \\
\bottomrule
\end{tabular}
\end{table}

\begin{table}[h]
\centering
\caption{Toy MDP RMSE by bin count $B$ ($\non = 100$, $K = 3$; mean over $10$ seeds).}
\label{tab:toy-bins}
\small
\begin{tabular}{lccccccc}
\toprule
Method & $B{=}4$ & $8$ & $16$ & $32$ & $64$ & $128$ & $256$ \\
\midrule
BC     & $0.617$ & $0.517$ & $0.473$ & $0.451$ & $0.440$ & $0.449$ & $0.467$ \\
\GBC{} & $\mathbf{0.378}$ & $\mathbf{0.374}$ & $\mathbf{0.372}$ & $\mathbf{0.364}$ & $\mathbf{0.368}$ & $\mathbf{0.379}$ & $\mathbf{0.400}$ \\
\bottomrule
\end{tabular}
\end{table}

\begin{table}[!ht]
\centering
\caption{Toy MDP RMSE by on-policy sample size $\non$ ($K = 3$, $B = 32$; mean $\pm$ SE over $10$ seeds).}
\label{tab:toy-non}
\small
\begin{tabular}{lcccc}
\toprule
Method & $\non{=}50$ & $\non{=}100$ & $\non{=}200$ & $\non{=}400$ \\
\midrule
Raw    & $0.429 \pm 0.005$ & $0.429 \pm 0.005$ & $0.429 \pm 0.005$ & $0.429 \pm 0.005$ \\
BC     & $0.452 \pm 0.005$ & $0.451 \pm 0.006$ & $0.452 \pm 0.005$ & $0.450 \pm 0.005$ \\
Direct & $0.507 \pm 0.022$ & $0.474 \pm 0.036$ & $0.405 \pm 0.017$ & $0.323 \pm 0.017$ \\
\GBC{} & $\mathbf{0.384 \pm 0.007}$ & $\mathbf{0.364 \pm 0.010}$ & $\mathbf{0.347 \pm 0.017}$ & $\mathbf{0.277 \pm 0.022}$ \\
\bottomrule
\end{tabular}
\end{table}

\paragraph{CRM: importance-weighted fitting of $\vhat$.}
We fit $\vhat$ by importance-weighted fitted value iteration instead of the unweighted fitting used for Table~\ref{tab:crm-headline}. The RMSE of \GBC{} is lower than that of Raw by $25.4 \pm 6.9\%$ (mean $\pm$ standard deviation over the four settings), compared with $26.2 \pm 6.0\%$ with unweighted fitting (Table~\ref{tab:crm-headline}).

\FloatBarrier
\paragraph{D4RL: TD and MC training of $\ehat$.}
Table~\ref{tab:d4rl-td-mc} compares TD training of $\ehat$ with MC training on the runs of Table~\ref{tab:d4rl-headline}. The MC training target for a state $S$ in $\Don$ is $G - \vhat(S)$, where $G$ is the discounted return from $S$ to the end of its $100$-step on-policy window. Direct (MC) and \GBC{} (MC) both use this $\ehat$.

\begin{table}[h]
\centering
\caption{D4RL RMSE with TD and MC training of $\ehat$ (mean $\pm$ SE over $10$ seeds). The rows for Raw and TD training are those of Table~\ref{tab:d4rl-headline}.}
\label{tab:d4rl-td-mc}
\small
\begin{tabular}{lccc}
\toprule
Method & HalfCheetah & Hopper & Walker2d \\
\midrule
Raw            & $25.66 \pm 0.81$ & $4.95 \pm 0.17$ & $7.20 \pm 0.23$ \\
Direct (TD)    & $23.43 \pm 0.75$ & $\mathbf{4.60 \pm 0.15}$ & $6.79 \pm 0.22$ \\
\GBC{} (TD)    & $22.98 \pm 0.68$ & $\mathbf{4.58 \pm 0.15}$ & $\mathbf{6.69 \pm 0.22}$ \\
Direct (MC)    & $\mathbf{19.81 \pm 0.45}$ & $5.86 \pm 0.19$ & $10.59 \pm 0.52$ \\
\GBC{} (MC)    & $22.98 \pm 0.83$ & $\mathbf{4.58 \pm 0.16}$ & $6.81 \pm 0.23$ \\
\bottomrule
\end{tabular}
\end{table}

\FloatBarrier
\subsection{KuaiRec-based recommender evaluation}\label{app:kuairec}

We also evaluate \GBC{} on a semi-synthetic recommender task in which the estimators observe only one-step transitions. It is built on the small matrix of KuaiRec \citep{gao2022kuairec}, in which almost every user--video pair has an observed watch ratio, which serves as the reward after scaling to $[0, 1]$. The policies and the session dynamics are constructed in the style of EasyRL4Rec \citep{yu2024easyrl4rec}. The state combines a low-rank user embedding with the categories of recently shown items, and a session lasts at most $30$ turns. The target policy samples from the $50$ items with the highest predicted rewards for the user, the behavior policy is an equal mixture of the target policy and the uniform policy over items, and $\gamma = 0.9$.

The estimators receive shuffled one-step transitions without episode identifiers or returns, so only TD training of $\ehat$ is possible. The behavior policy generates $5 \times 10^4$ transitions to fit $\vhat$ and another $5 \times 10^4$ that form $\Doff$, and $\Don$ consists of $5 \times 10^3$ target-policy transitions. The base predictor is a linear model fit to the value of the behavior policy by fitted value iteration, as on Toy; the error model is a random forest; and the calibration follows Appendix~\ref{app:impl}. The ground truth $\vzero$ is estimated at $250$ evaluation states by $500$ target-policy rollouts each.

Table~\ref{tab:kuairec} reports RMSE. \GBC{} is more accurate than BC and Direct, with paired differences (BC or Direct minus \GBC{}) of $0.009 \pm 0.002$ and $0.140 \pm 0.003$.

\begin{table}[h]
\centering
\caption{KuaiRec-based evaluation: RMSE (mean $\pm$ SE over $10$ seeds).}
\label{tab:kuairec}
\small
\begin{tabular}{lc}
\toprule
Method & RMSE \\
\midrule
Raw           & $0.593 \pm 0.008$ \\
BC            & $0.376 \pm 0.006$ \\
Direct        & $0.507 \pm 0.008$ \\
\GBC{} (ours) & $\mathbf{0.368 \pm 0.007}$ \\
\bottomrule
\end{tabular}
\end{table}

\section{Extended Related Work}\label{app:related}

\subsection{Bellman calibration and direct predecessors}\label{app:rw-bc}

The direct predecessor of this paper is the Bellman calibration framework of \citet{vanderlaan2025bc}, which calibrates a value predictor $\vhat$ by fitting a one-dimensional histogram regression along the level sets of $\vhat$ against a doubly robust Bellman target. Their main theoretical results are a calibration-refinement decomposition that exposes the within-level-set variation of the prediction error as the refinement error, a histogram-binning calibration error rate, and a finite-sample value-estimation bound. \citet{vanderlaan2025bc} also note that the calibration step can be run on a limited amount of on-policy data; \GBC{} instead uses the on-policy sample to learn the conditioning axis. Iterated calibration in MDPs also appears in \citet{vanderlaan2025sdrl}, who calibrate an initial $Q$-function estimate by fitted $Q$-iteration with isotonic regression to debias plug-in estimators of policy values, whereas BC and \GBC{} calibrate the state-value predictor itself. The companion paper of \citet{vanderlaan2025swfqe} reweights the fitted Q evaluation objective with a stationary density ratio, which removes the Bellman completeness requirement from training; it changes the predictor, whereas BC and \GBC{} adjust the outputs of a fixed predictor.

\subsection{Calibration foundations}\label{app:rw-cal}

The conceptual roots of the calibration-refinement decomposition trace to \citet{murphy1973vector} and \citet{degroot1983comparison}, who decomposed the Brier score of a probability forecaster into a calibration component and a refinement component. The Bellman calibration framework adapts this decomposition to value prediction. The calibration-with-nuisance program of \citet{vanderlaan2023ic} introduced causal isotonic calibration for treatment-effect predictors using doubly robust pseudo-outcomes, and \citet{whitehouse2024orthogonal} generalized this construction via Neyman orthogonality to lift any calibration algorithm into the setting of nuisance-dependent targets. Bellman calibration carries this lineage to MDPs, with a doubly robust Bellman target in place of a static treatment effect.

\subsection{Off-policy evaluation}\label{app:rw-ope}

Our DR Bellman target belongs to the doubly robust line of off-policy evaluation, from the contextual-bandit estimator of \citet{dudik2011doubly} to the sequential estimator of \citet{jiang2016dr} and the double reinforcement learning formulation of \citet{kallus2020drl}, which combines a stationary density ratio \citep{liu2018curse} with a $Q$-function nuisance. Fitted Q evaluation \citep{le2019batch} is a standard choice of base predictor. The product of two nuisance errors in the term $\termN$ of Theorem~\ref{thm:est-error} goes back to the double/debiased machine learning framework of \citet{chernozhukov2018ddml}.

\subsection{Bellman completeness barriers}\label{app:rw-completeness}

The Bellman completeness assumption originates in the analysis of fitted value iteration by \citet{munos2008fitted}, where it is needed to control the propagation of approximation error across iterations. \citet{chen2019information} study the roles of completeness and concentrability in batch reinforcement learning. \citet{foster2022barriers} prove that realizability and concentrability alone are insufficient for sample-efficient offline value-function approximation, and \citet{wang2021statistical} establish a lower bound for offline RL with linear function approximation showing that completeness-style structural conditions are essential for tractable learning. \citet{xie2021batch} provide a partial positive result via a tournament procedure under realizability alone. The calibration guarantees of BC and \GBC{} do not assume completeness, and neither does the MC refinement bound of Proposition~\ref{prop:mc-refinement}; the TD refinement bound of Proposition~\ref{prop:td1-refinement} instead contains the inherent Bellman error of $\Ppi$ on $\sigma(\ehat)$ (Definition~\ref{def:ibe-ehat}).

\subsection{Hybrid and semi-offline reinforcement learning}\label{app:rw-hybrid}

Hybrid reinforcement learning also uses a small amount of online interaction alongside a large offline dataset, but targets policy learning rather than the calibration of a given value predictor. Examples include system identification for model-based RL \citep{ross2012agnostic}, hybrid Q-learning \citep{song2023hybrid}, reward-agnostic analyses in the tabular setting \citep{wagenmaker2023reward}, offline pre-training followed by online fine-tuning \citep{nakamoto2023calql}, and an offline-enhanced policy-gradient method that avoids Bellman completeness \citep{zhou2023offlinepg}.

\subsection{Low-dimensional adaptation in nonparametric regression}\label{app:rw-lowdim}

The intrinsic-dimension rate $\non^{-\beta/(2\beta+k)}$ that drives Theorem~\ref{thm:rate-transfer} is the classical Stone rate for $\beta$-H\"older regression on $k$ effective dimensions, established as a minimax-optimal lower bound by \citet{stone1980optimal} and as a global rate by \citet{stone1982optimal}. Assumption~\ref{ass:LS2} takes such a rate as given for the chosen error-model estimator rather than deriving it in our setting. Rates of this form have been established, each under its own conditions, for sparse ReLU networks under compositional H\"older structure \citep{schmidthieber2020nonparametric} and manifold inputs \citep{schmidthieber2019manifold}, deep networks under hierarchical interaction models \citep{bauer2019deeplearning} and under low Minkowski dimension of the input support \citep{nakada2020adaptive}, and kernel ridge regression with effective-dimension control \citep{caponnetto2007optimal}. Because Theorem~\ref{thm:rate-transfer} takes the rate as an input, a better rate for $\ehat$ carries over directly to the refinement bound.

\end{document}